\documentclass[10pt,twocolumn,letterpaper]{article}

\usepackage{listings}
\usepackage{colortbl}
\usepackage{array} 
\usepackage{float}
\definecolor{mypink}{RGB}{255,200,190} 
\definecolor{myblue}{RGB}{203,203,254} 
\usepackage{capt-of} 
\usepackage{graphicx}
\usepackage{cuted}
\usepackage{times}

\usepackage{style}
\usepackage{adjustbox}
\usepackage{graphicx}
\usepackage{booktabs}
\usepackage{multirow}
\usepackage{multicol}
\usepackage{array}
\usepackage[dvipsnames,table]{xcolor}
\usepackage{caption}
\usepackage{float}
\usepackage{algorithm}
\usepackage{algorithmic}
\usepackage{bbm}

\definecolor{styleblue}{rgb}{0.21,0.49,0.74}
\usepackage[pagebackref,breaklinks,colorlinks,allcolors=styleblue]{hyperref}

\title{FlowDirector: Training-Free Flow Steering for Precise Text-to-Video Editing}

\author{Guangzhao Li$^{*1,2}$ \:\: Yanming Yang$^{1}$ \:\: Chenxi Song$^{1}$ \:\: Xiaohong Liu$^{3}$ \:\: Chi Zhang$^{\dagger1}$ \\
$^{1}$AGI Lab, Westlake University \quad\quad  $^{2}$Central South University \quad\quad  $^{3}$Shanghai Jiao Tong University.  \\ \\
\url{https://flowdirector-edit.github.io}
}

\begin{document}
\maketitle

\begin{strip}
\begin{minipage}{\textwidth}\centering
\vspace{-30pt}
\includegraphics[width=0.88\linewidth]{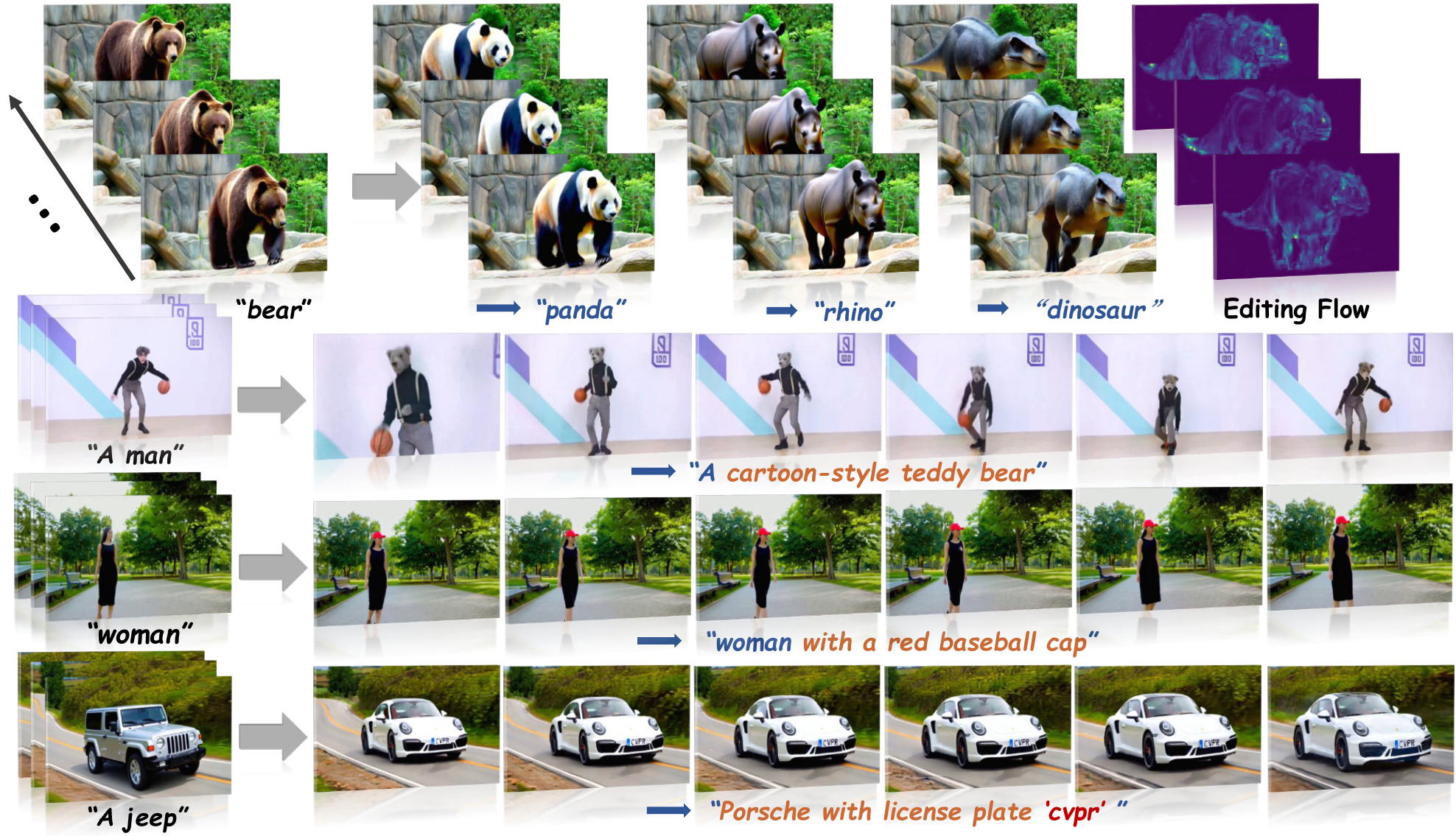}
\captionof{figure}{\textbf{Editing results.} Our method uses a source video and prompt to generate precise, semantic edits while preserving background content and motion, yielding high visual fidelity and temporal coherence results.}
\label{fig:teaser}
\end{minipage}
\end{strip}

{
  \renewcommand{\thefootnote}%
    {\fnsymbol{footnote}}
 \footnotetext[1]{This work was done during Guangzhao Li’s visit at AGI Lab, Westlake
University.\\
\hspace*{1.35em}$^{\dagger}$ denotes corresponding author.
  }
}

\begin{abstract}
Text-driven video editing aims to modify video content based on natural language instructions. While recent training-free methods have leveraged pretrained diffusion models, they often rely on an inversion-editing paradigm. This paradigm maps the video to a latent space before editing. However, the inversion process is not perfectly accurate, often compromising appearance fidelity and motion consistency.
To address this, we introduce FlowDirector, a novel training-free and inversion-free video editing framework. Our framework models the editing process as a direct evolution in the data space. It guides the video to transition smoothly along its inherent spatio-temporal manifold using an ordinary differential equation (ODE), thereby avoiding the inaccurate inversion step.
From this foundation, we introduce three flow correction strategies for appearance, motion, and stability: 1) Direction-aware flow correction amplifies components that oppose the source direction and removes irrelevant terms, breaking conservative streamlines and enabling stronger structural and textural changes. 2) Motion-appearance decoupling optimizes motion agreement as an energy term at each timestep, significantly improving consistency and motion transfer. 3) Differential averaging guidance strategy leverages differences among multiple candidate flows to approximate a low variance regime at low cost, suppressing artifacts and stabilizing the trajectory. 
Extensive experiments across various editing tasks and benchmarks demonstrate that FlowDirector achieves state-of-the-art performance in instruction following, temporal consistency, and background preservation, establishing an efficient new paradigm for coherent video editing without inversion.
\end{abstract}

\section{Introduction}
\label{sec:intro}

Text-driven video editing aims to modify video content based on natural language instructions, offering a powerful and intuitive interface for content creation and manipulation.
The rapid development of this field has been largely fueled by recent advances in generative artificial intelligence, particularly the emergence of large-scale pre-trained diffusion models~\cite{ddpm, song2020score0based} such as Stable Diffusion~\cite{rombach2021high0resolution, podell2023sdxl0}. These models are trained on extensive image-text datasets~\cite{lin2014microsoft, changpinyo2021conceptual, LAION} and encode rich visual and semantic priors, enabling them to synthesize high-quality and semantically aligned images from textual descriptions. Building upon these capabilities, recent research efforts~\cite{turn-a-video, video-p2p, edit-a-video, fatezero, text2video-zero, tokenflow, flatten, ground-a-video, rave} have extended diffusion models from static image generation to dynamic video domains.
Such extensions have given rise to training-free video editing~\cite{fatezero, tokenflow, flatten, rave, videodirector, StableV2V, adaflow, reattentional}, which directly manipulates video frames under textual guidance by reusing the knowledge embedded in pre-trained diffusion backbones.

Despite promising progress, text-driven video editing presents unique and significant challenges that distinguish it from image editing. Videos are inherently high-dimensional sequences with rich temporal dynamics, requiring coherence not only within each frame but also across time.
Most existing training-free methods~\cite{fatezero, tokenflow, flatten, rave, videodirector} rely on inversion-based~\cite{ddim-inversion} paradigm, mapping the input video to a latent trajectory. This approach works effectively for images, but it presents significant challenges for video.
Inverting an entire video sequence coherently requires generating a temporally smooth latent trajectory, which existing image-based inversion techniques are not designed to produce. In practice, producing such a coherent latent trajectory is difficult. For instance, \cite{videodirector} shows that the reconstructed results have significant differences in appearance and motion compared to the source video.
Small per-frame inversion errors accumulate and cause temporal drift and flicker. Attention misalignment across frames disrupts layout and identity, and mixing motion and appearance leads to style inconsistency and unrealistic motion. Together, these issues substantially degrade edit quality.

In this work, we explore a novel inversion-free framework for text-driven video editing that overcomes the fundamental limitations of prior methods. Inspired by recent progress in flow-based inversion-free image editing~\cite{infedit, flowedit}, we develop an editing paradigm that models the transformation from the original video to the edited result as a direct evolution in data space, governed by a learned Ordinary Differential Equation (ODE).
Unlike traditional approaches that attempt to recover precise latent representations for each frame through inversion, the inversion-free paradigm constructs a smooth transformation path that guides the video along its native spatiotemporal manifold, achieving significant improvements in editing fidelity and temporal consistency compared to traditional inversion methods. 
However, a straightforward application of this paradigm to video editing still presents significant challenges. First, the temporal dimension introduces substantially richer appearance and structural information across frames, making it more difficult to achieve drastic semantic transformations while preserving spatiotemporal plausibility. 
Second, the lack of explicit strong constraints leads to severe motion distortion and drift in the editing results. Besides, the compounded sampling noise across frames makes the system exceptionally sensitive to perturbations, undermining stability throughout the video sequence.
To tackle these challenges, we develop three complementary, training-free flow correction strategies that jointly optimize appearance preservation, motion stability, and spatiotemporal coherence, yielding significantly improved editing quality and robustness.

Standard inversion-free methods often skip early high-noise diffusion steps to maintain structural consistency.  This is in contrast to traditional inversion-based paradigms, which actively disrupt these strong structural constraints by mapping data to noise, thereby creating the necessary latitude for significant manipulation. As a result, inversion-free method retains strong content information from the source video, creating a ``semantic gravitational pull'' that resists drastic alterations. In this regime, the standard editing flow is frequently overwhelmed by the dominant source priors, resulting in overly conservative edits. To overcome this inherent resistance, we introduce \textbf{Direction-Aware Flow Correction} (DA-FC). Our key insight is that the raw editing flow contains opposing forces: components aligned with the source that cause redundant drift, and components opposing the source that drive actual semantic change. DA-FC mathematically decouples these forces, actively suppressing the aligned components while significantly amplifying the opposing ones. This strategy effectively breaks the semantic dominance of the source video, enabling profound structural transformations without sacrificing the stability benefits of the inversion-free paradigm.

Another critical challenge in direct ODE editing is the entanglement of motion and appearance flows. In many editing scenarios, we aim for conflicting goals: drastic appearance transformations (\eg, turning a person into a bear) must coexist with strict motion preservation (\eg, maintaining the exact trajectory of a dribbled basketball). Unlike inversion-based approaches that can inject explicit structural guidance (\eg, cross-attention maps) from the source video, inversion-free methods lack such isolated constraints. Consequently, during complex dynamics like occlusions, errors in motion can accumulate and be misinterpreted by the model as necessary appearance changes, leading to severe distortion (\eg, objects shifting incorrectly in depth). Simply enforcing similarity to the source video to fix this is counterproductive, as it would actively revert the desired appearance edits. To resolve this conflict, we propose \textbf{Motion-Appearance Decoupling Flow Correction} (MAD-FC). The core motivation is to establish an orthogonal control plane: by mathematically isolating ``pure motion'' features from static appearance information, we can formulate an energy function that strictly penalizes motion deviations while remaining completely agnostic to the intended semantic transformations. This allows for continuous, ``safe'' rectification of temporal drift without compromising the magnitude of the edit.

Finally, direct ODE editing faces the challenge of high-variance velocity estimates, where single-sample noise introduces directional jitter that accumulates into severe temporal flickering. While brute-force averaging~\cite{flowedit} can mitigate this in image editing, it is computationally intractable for the high-dimensional video domain. We propose \textbf{Differential Averaging Guidance} (DAG) to transcend this limitation. Instead of passively relying on massive sampling to dilute noise—which is too costly for video—DAG adopts an active navigation strategy. By explicitly contrasting a high-quality consensus estimate against a baseline of high-variance outliers, we extract a ``noise drift'' signal. DAG then uses this signal not just to average out errors, but to actively steer the trajectory away from these noisy deviations, efficiently locking the editing process onto a stable, low-variance manifold with minimal computational overhead.

These three correction strategies, integrated with our direct editing ODE, constitute our framework, FlowDirector. Extensive experiments on standard benchmarks demonstrate that our method consistently outperforms existing training-free baseline methods across multiple key dimensions, including adherence to editing instructions, semantic consistency, motion consistency, and background preservation. Overall, our contributions are summarized as follows:
\begin{itemize}
     
\item We propose \textbf{FlowDirector}, a novel training-free video editing framework. It employs a novel inversion-free paradigm and can perform various video editing tasks, providing new understanding and insights for effective video editing.
\item We propose three training-free flow correction strategies: direction-aware flow correction, motion-appearance decoupling flow correction, and differential averaging guidance, which significantly improve appearance editing, motion preservation, and editing stability.
\item Extensive experiments demonstrate that our method achieves state-of-the-art performance across various editing scenarios and benchmarks. The code will be publicly released to facilitate further research.

\end{itemize}

\begin{figure*}[tb]
  \centering
  \includegraphics[width=0.88\textwidth]{figures/method_overview.pdf}
  \vspace{-2mm}
  \caption{\textbf{We compare inversion-based methods, FlowDirector (Direct ODE), and the full FlowDirector.} 
\textbf{Inversion-based methods} first map the source video into a Gaussian latent space and then use this inverted state as the starting point for the subsequent editing process. 
\textbf{FlowDirector (Direct ODE)} instead constructs source-side and target-side states at each timestep, estimates their velocity fields, and uses the difference between them as an editing flow that drives video editing directly in data space. 
Building on this formulation, the full \textbf{FlowDirector} further corrects the editing flow at every timestep, yielding a shorter and more efficient editing trajectory and substantially improving the final editing quality.}
  \label{fig:qr}
\vspace{-4mm}
\end{figure*}

\section{Related Works}
\label{sec:intro}

\textbf{Text-to-Image Editing} Diffusion model-based~\cite{ddpm, song2020score0based, flowmatching} image editing is primarily classified into two paradigms. Inversion-based methods~\cite{p2p, ddim-inversion, imagic, InstructPix2Pix, sdedit, TurboEdit, DiffusionCLIP} use an invert-then-denoise strategy: source images are mapped to latent noise via deterministic sampling (\eg, DDIM Inversion~\cite{ddim, ddim-inversion}), with editing by manipulating internal representations during denoising. Their limitation is that inversion approximation errors degrade reconstruction quality and editing fidelity. To overcome this, inversion-free methods emerged. InfEdit~\cite{infedit} uses Denoising Diffusion Consistent Models for virtual inversion. FlowEdit~\cite{flowedit} uses Flow Matching Models to construct a direct ordinary differential equation (ODE) trajectory between the source and target distributions, bypassing inversion. FlowAlign~\cite{flowalign} improves editing quality by optimizing the sampling path.

\textbf{Text-to-Video Editing} is primarily categorized by the underlying generative model. Early approaches adapted T2I models~\cite{rombach2021high0resolution, imagen, sd3.5, podell2023sdxl0}, facing temporal consistency challenges due to a lack of temporal understanding. Some are zero-shot (FateZero~\cite{fatezero}, TokenFlow~\cite{tokenflow}, Flatten~\cite{flatten}, RAVE~\cite{rave}), others require fine-tuning (Tune-A-Video~\cite{turn-a-video}, Video-P2P~\cite{video-p2p}). These T2I-based methods often struggle with temporal consistency. Recent work leverages native T2V models~\cite{make-a-video, cogvideo, vdm, svd, animatediff, hunyuanvideo, wan, ModelScope, MagicVideo-v2, VideoCrafter1} and their learned spatiotemporal priors for improved temporal consistency (\eg, VideoSwap~\cite{videoswap}, VideoDirector~\cite{videodirector}). However, all the aforementioned video editing methods are inversion-based. Similar to image editing, their inherent limitations restrict substantial visual transformations, resulting in insufficient editing precision. Furthermore, the accumulation of temporal errors in the two stages of inversion and denoising often causes motion distortion or drift.

\begin{figure}[tb]
  \centering
  \includegraphics[width=0.88\linewidth]{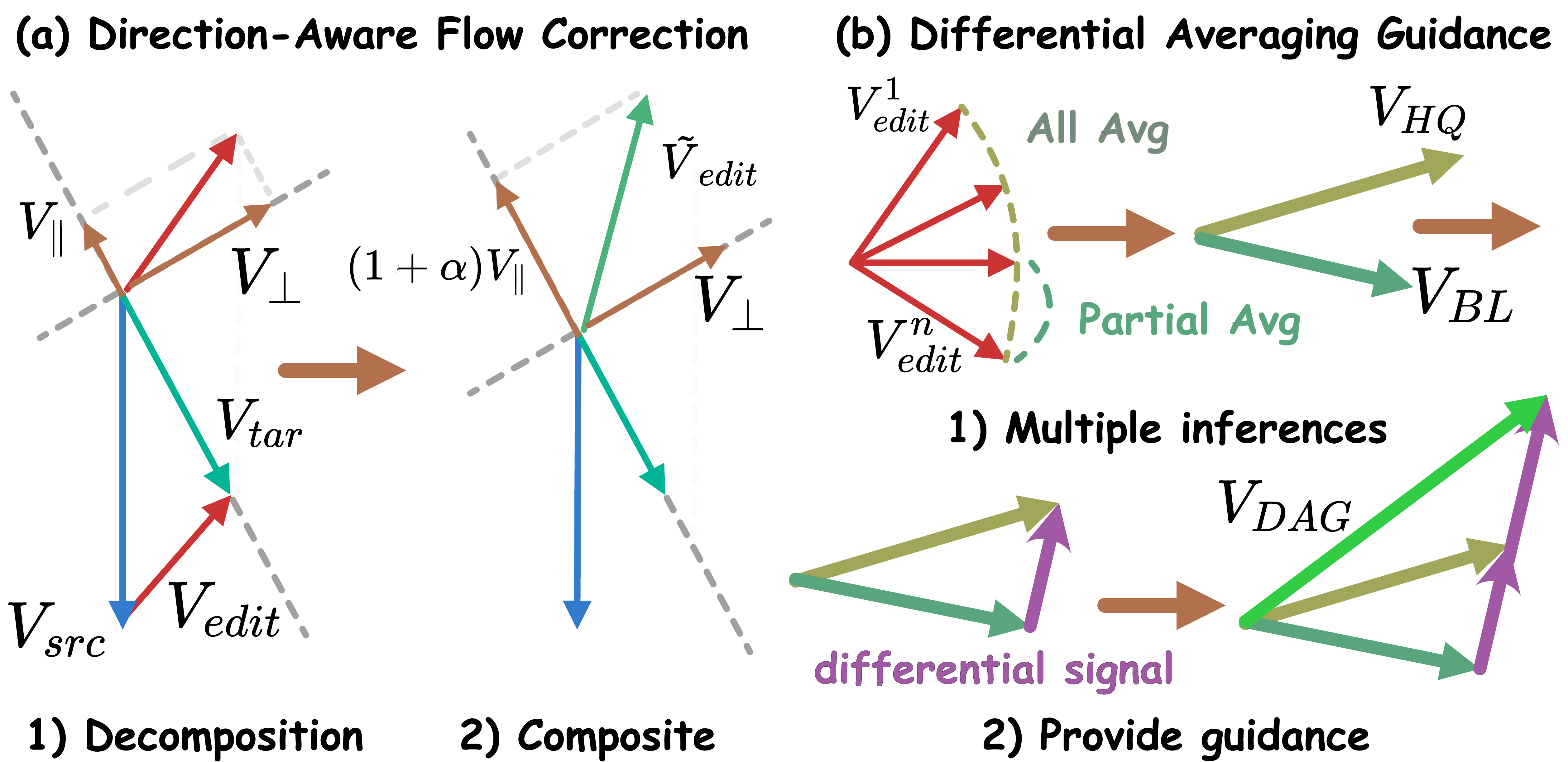}
\vspace{-1mm}
  \caption{\textbf{Illustration of proposed Direction-Aware Flow Correction (DA-FC, a) and Differential Averaging Guidance (DAG, b).} DA-FC strengthens editing through orthogonal decomposition of the editing flow, amplifying parallel components that oppose the source semantics while eliminating co-directional components. (a) shows the case of amplifying the inverse component. DAG guides the editing flow toward a stable state by generating differential signals.}
  \label{fig:qr}
\vspace{-4mm}
\end{figure}

\section{Method}
\label{sec:method}

\subsection{Preliminary}
\label{sec:preliminary}

Our framework builds on two standard components in flow-based generative modeling.

\textbf{Flow Matching.} Flow Matching learns a time-dependent vector field that transports a simple prior (\eg, $\mathcal{N}(0,I)$) to the data distribution. A neural network $v_\theta(x,t)$ is trained by minimizing
\begin{equation} \label{eq:fm_loss_en_final}
\mathcal{L}_{\text{FM}}
= \mathbb{E}\!\left[ \,\|v_\theta(x,t) - v_t(x\,|\,x_1)\|^2 \right],
\end{equation}
where $t\!\sim\!\mathcal{U}(0,1)$, $x_1\!\sim\!p_1$, $x\!\sim\!p_t(\cdot\,|\,x_1)$, $p_t(x\,|\,x_1)$ is a prescribed path from $p_0$ to $x_1$, and $v_t(x\,|\,x_1)$ is its oracle velocity.

\textbf{Rectified Flow.} Rectified Flow~\cite{Rectified-Flow} specializes Flow Matching to linear paths between $p_0$ and $p_1$. Given $x_0 \sim p_0$ and $x_1 \sim p_1$, it considers $x_t = (1-t)x_0 + t x_1$ and learns
\begin{equation} \label{eq:rf_loss}
\mathcal{L}_{\text{RF}}
= \mathbb{E}\!\left[ \,\|v_\theta(x_t,t) - (x_1 - x_0)\|^2 \right],
\end{equation}
where $t\!\sim\!\mathcal{U}(0,1)$, $x_0\!\sim\!p_0$, $x_1\!\sim\!p_1$. This linearization simplifies the oracle velocity and enables efficient, stable ODE integration driven by $v_\theta$.

\subsection{Editing Flow Generation}
\label{sec:editing-flow-generation}

To avoid the inversion stage that often harms visual fidelity and motion consistency, we directly construct an editing trajectory in the data space using a pre-trained text-to-video (T2V) model $v_\theta$. Given a source video $X_{\text{src}}$ with prompt $c_{\text{src}}$ (or an automatically inferred description by LLM if $c_{\text{src}}$ is not provided) and a target prompt $c_{\text{tar}}$, we define the editing state at time $t \in [0,1]$ as $Z_t^{\text{edit}}$. Following FlowEdit~\cite{flowedit}, this state is constructed via the unified editing equation:
\begin{equation} \label{eq:edit_rephrased}
  Z_t^{\text{edit}} = X_{\text{src}} - Z_t^{\text{src}} + Z_t^{\text{tar}},
\end{equation}
where $Z_t^{\text{src}}$ and $Z_t^{\text{tar}}$ are the perturbed source and target states at time $t$. The evolution of $Z_t^{\text{edit}}$ is governed by an ODE whose velocity field is the difference between target and source velocities:
\begin{equation} \label{eq:video_flowedit_ode_main_revised}
  \frac{dZ_t^{\text{edit}}}{dt} = V_{\text{edit}}(t)
  = v_\theta(Z_t^{\text{tar}}, t, c_{\text{tar}})
  - v_\theta(Z_t^{\text{src}}, t, c_{\text{src}}).
\end{equation}
The trajectory starts from the source video $Z_{1}^{\text{edit}} = X_{\text{src}}$ and converges to the edited result $Z_0^{\text{edit}}$ as $t \to 0$.

To compute $V_{\text{edit}}(t)$, we need the intermediate states $Z_t^{\text{src}}$ and $Z_t^{\text{tar}}$. Following Rectified Flow~\cite{Rectified-Flow}, we obtain the source state by linear interpolation
$Z_t^{\text{src}} = (1 - t) X_{\text{src}} + t N_t$, where $N_t \sim \mathcal{N}(0, I)$.
Since the target video is not observed, we derive
$Z_t^{\text{tar}}$ by rearranging Eq.~\eqref{eq:edit_rephrased}:
$Z_t^{\text{tar}} = Z_t^{\text{edit}} + Z_t^{\text{src}} - X_{\text{src}}$.
This construction ensures that $Z_t^{\text{src}}$ and $Z_t^{\text{tar}}$ share the same noise $N_t$, so noise and other random perturbations cancel in the difference $V_{\text{edit}}(t)$, and the resulting editing flow primarily captures the semantic change induced by the prompts.

\begin{figure*}[tb]
  \centering
  \includegraphics[width=0.9\textwidth]{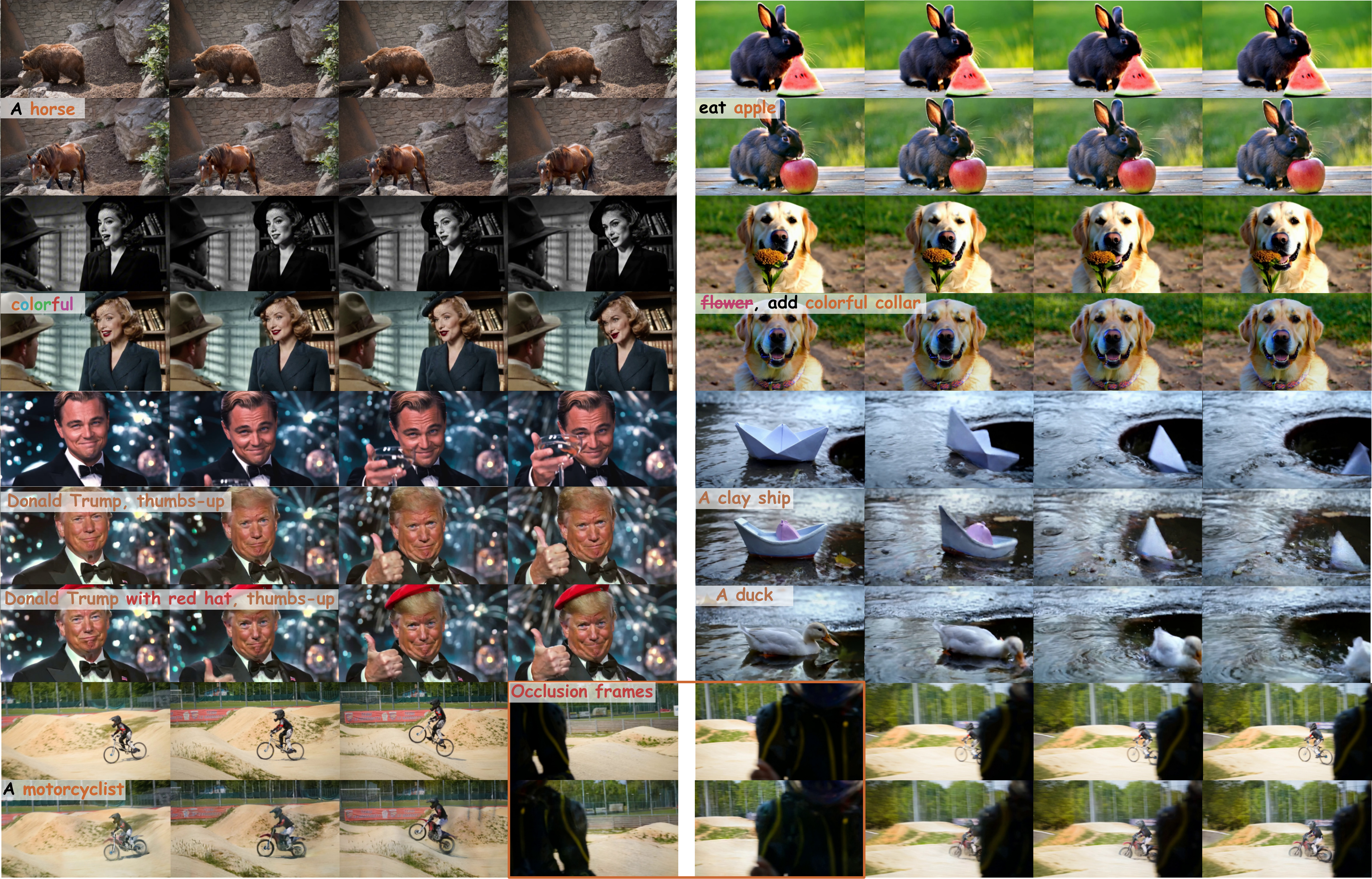}
  \vspace{-2mm}
  \caption{\textbf{Qualitative results of FlowDirector.} Our method can perform various types of editing tasks. The results demonstrating high fidelity to prompts, preservation of unedited regions and motion, temporal coherence, and visual plausibility.}
  \label{fig:qr}
\vspace{-4mm}
\end{figure*}

\subsection{Direction-Aware Flow Correction}
\label{sec:direction-aware-flow-correction}

To promote thorough edits while improving stability, we introduce \emph{Direction-Aware Flow Correction} (DA-FC). At each timestep we obtain the source and target velocities
$V_{\text{src}} = v_\theta(Z_t^{\text{src}}, t, c_{\text{src}})$ and
$V_{\text{tar}} = v_\theta(Z_t^{\text{tar}}, t, c_{\text{tar}})$, and define the editing velocity
$V_{\text{edit}} = V_{\text{tar}} - V_{\text{src}}$ (we omit $t$ below for brevity).
We perform a token-wise orthogonal decomposition of $V_{\text{edit}}$ along $V_{\text{src}}$:
\begin{equation}
\label{eq:dircorr-decomp}
V_{\parallel} = \frac{\langle V_{\text{edit}},\, V_{\text{src}}\rangle}{\|V_{\text{src}}\|^2 + \varepsilon}\, V_{\text{src}}, 
\quad 
V_{\perp} = V_{\text{edit}} - V_{\parallel},
\end{equation}
where $\langle\cdot,\cdot\rangle$ and $\|\cdot\|$ denote the Euclidean inner product and $\ell_2$ norm, and $\varepsilon>0$ is a small constant for numerical stability.
We then apply the direction-aware correction:
\begin{equation}
\label{eq:dircorr-rule}
\tilde{V}_{\text{edit}} =
\begin{cases}
V_{\perp}, & \text{if } \langle V_{\parallel}, V_{\text{src}}\rangle \ge 0,\\[2pt]
(1+\alpha)\,V_{\parallel} + V_{\perp}, & \text{if } \langle V_{\parallel}, V_{\text{src}}\rangle < 0,
\end{cases}
\end{equation}
where $\alpha>0$ is an amplification factor. Thus, components aligned with the source direction are suppressed to reduce redundant drift, while anti-aligned components are amplified to encourage meaningful structural changes, yielding a more decisive and stable editing trajectory.

The corrected flow still acts on the full spatial extent and may alter irrelevant regions. To localize edits, we derive an editing mask from cross-attention maps of the source and target prompts. We first aggregate cross-attention maps over heads and token dimensions, apply 2D average pooling and normalization per frame, and then threshold by the mean to obtain $\mathbf{M}_{\mathrm{src}}$. The same procedure on target-prompt keywords yields $\mathbf{M}_{\mathrm{tar}}$, and we define $\mathbf{M}:=\mathbf{M}_{\mathrm{src}}\cup\mathbf{M}_{\mathrm{tar}}$. To allow smooth transitions between edited regions and background, we soften this mask by applying a Euclidean distance transform $d$ to the background and an exponential decay:
\begin{equation}
\widetilde{\mathbf{M}}_{c,t}(x,y)
=\mathbf{M}_{c,t}(x,y)+\bigl(1-\mathbf{M}_{c,t}(x,y)\bigr)\,e^{-\lambda\, d_{c,t}(x,y)},
\end{equation}
where $(x,y)$ indexes spatial locations, $c$ and $t$ denote channel and time, and $\lambda>0$ controls the decay rate.
Finally, we apply the softened mask to the corrected editing flow:
\begin{equation} \label{eq:SAFC}
\hat{V}_{\text{edit}} \;=\; \tilde{V}_{\text{edit}} \odot \widetilde{\mathbf{M}},
\end{equation}
where $\odot$ denotes element-wise multiplication.

\subsection{Motion-Appearance Decoupling Correction}
\label{sec:motion-appearance-decoupling-flow-correction}

To maintain motion consistency in an inversion-free setting, we introduce \emph{Motion-Appearance Decoupling Flow Correction} (MAD-FC), which enforces a dynamic motion constraint at each denoising step. At time $t$, we first estimate the noiseless source and target states using the Tweedie estimator,
$Z^{\mathrm{src}}_0 = Z^{\mathrm{src}}_t - t\,V_{\mathrm{src}}$ and 
$Z^{\mathrm{tar}}_0 = Z^{\mathrm{tar}}_t - t\,V_{\mathrm{tar}}$,
and compute temporal averages on each side to obtain anchor frames $(A^{\mathrm{src}}_h, A^{\mathrm{tar}}_h)$ as appearance references. Motion representations are then defined by subtracting the anchors,
$G_{\mathrm{src}} = Z^{\mathrm{src}}_0 - A^{\mathrm{src}}_h$ and 
$G_{\mathrm{tar}} = Z^{\mathrm{tar}}_0 - A^{\mathrm{tar}}_h$,
which suppress static appearance and emphasize motion. Finally, we measure motion disagreement between source and target using the energy:
\begin{equation}\label{eq:madfc-energy}
J_t(Z) = \tfrac{1}{2}\,\|G_{\mathrm{tar}} - G_{\mathrm{src}}\|_2^2.
\end{equation}

After the standard ODE update $Z^{\mathrm{edit}}_t \leftarrow Z^{\mathrm{edit}}_t + \Delta t\,V_{\mathrm{edit}}$, we further minimize $J_t$ with respect to $Z_t$ by performing a single gradient-descent step. Using the first-order approximation $\nabla_{Z_t} J_t \approx G_{\mathrm{tar}} - G_{\mathrm{src}}$, we update the edited state in the negative gradient direction:
\begin{equation}\label{eq:madfc-update}
Z^{\mathrm{edit}}_t \leftarrow Z^{\mathrm{edit}}_t - \zeta\,(G_{\mathrm{tar}} - G_{\mathrm{src}}).
\end{equation}
Rewriting this in terms of $Z_0$ and the anchors yields the practical update rule:
\begin{equation}\label{eq:madfc-expanded}
Z^{\mathrm{edit}}_t \leftarrow Z^{\mathrm{edit}}_t
- \zeta\Big[
\underbrace{(Z^{\mathrm{tar}}_0 - Z^{\mathrm{src}}_0)}_{\text{motion}}
- \phi\,\underbrace{(A^{\mathrm{tar}}_h - A^{\mathrm{src}}_h)}_{\text{appearance}}
\Big].
\end{equation}
Here $\Delta t$ is the ODE integration step size, $\zeta>0$ controls the overall correction strength, and $\phi\in[0,1]$ modulates how strongly appearance alignment via the anchors is enforced.

In effect, MAD-FC reduces the motion energy between source and target at every timestep, transferring source motion to the edited video while allowing a tunable trade-off between appearance correction and motion preservation through $\phi$. This yields stable edits that remain consistent even under occlusions and complex dynamics.

\subsection{Differential Averaging Guidance}
\label{sec:differential-averaging-guidance}

To stabilize the editing trajectory and reduce variance, we introduce \emph{Differential Averaging Guidance} (DAG). The key idea is to form a high-quality velocity estimate by averaging multiple stochastic editing flows, and to use their discrepancy from a conservative baseline as a guidance signal.
At each sampling step $t$, we compute a high-quality estimate $V_{\text{HQ}}$ by averaging $L_{\text{HQ}}$ velocity fields obtained under different noise samples:
\begin{equation}\label{eq:10}
V_{\text{HQ}}(Z_t^{\text{edit}}, t)\;=\;\frac{1}{L_{\text{HQ}}}\sum_{\ell=1}^{L_{\text{HQ}}}{V}_{\text{edit}}^{(\ell)}(Z_t^{\text{tar}},Z_t^{\text{src}},t,c^{\text{tar}},c^{\text{src}}).
\end{equation}
From the same pool, we construct a baseline estimate $V_{\text{BL}}$ by selecting the $K$ candidates with the lowest cosine similarity to $V_{\text{HQ}}$ and averaging them:
\begin{equation}
V_{\text{BL}}(Z_t^{\text{edit}}, t)\;=\;\frac{1}{K}\sum_{i\in\mathcal{I}_K}{V}_{\text{edit}}^{(i)}(\cdot),
\end{equation}
where $\mathcal{I}_K$ indexes the $K$ least similar velocity fields, $K\!\le\!L_{\text{HQ}}$, and $(\cdot)$ denotes the same arguments as in Eq.~\ref{eq:10}. The differential signal
$\bar{D}\;=\;V_{\text{HQ}}-V_{\text{BL}}$
is then used to guide the final velocity:
\begin{equation}\label{eq:14}
V_{\text{DAG}}\;=\;V_{\text{HQ}} + w\,\bar{D},
\end{equation}
with $w>0$ controlling the guidance strength.

In flow-matching generative models, large shift values create wide step intervals late in sampling. Different noise draws then perturb the editing trajectory and cause texture flicker. We fix a single noise sample for the late denoising phase. After a chosen cutoff timestep, we stop resampling and reuse this noise for all remaining steps. This stabilizes textures and other fine details.

\begin{figure*}[ht]
  \centering
  \includegraphics[width=0.88\textwidth]{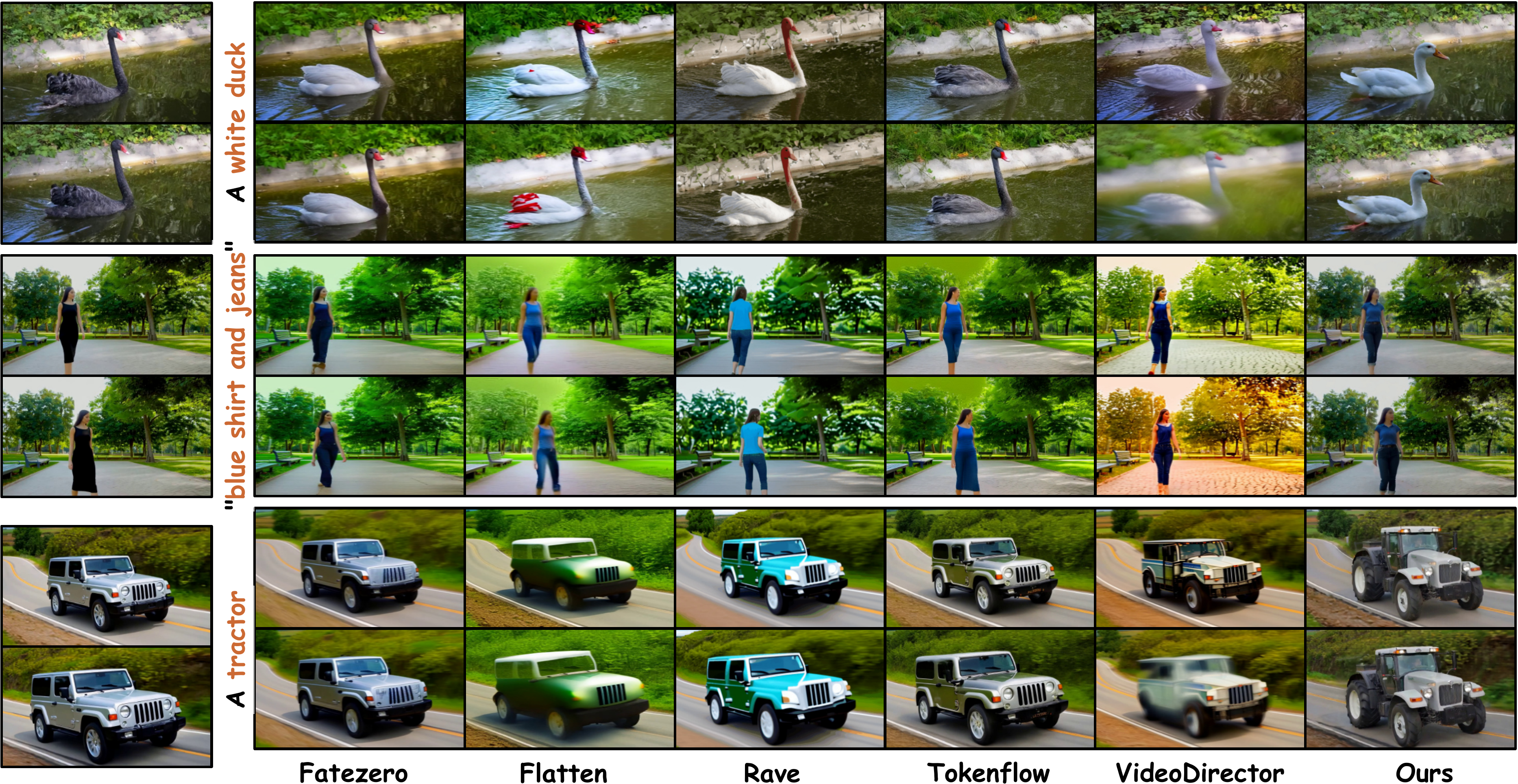}
  \vspace{-1mm}
  \caption{\textbf{Qualitative comparison.} Our method outperforms previous methods across diverse editing tasks, demonstrating superior visual quality and temporal consistency. Best viewed zoomed-in.}
  \label{fig:compare}
\vspace{-1mm}
\end{figure*}

\begin{table*}[htb]
    \centering
    \caption{
    \textbf{Quantitative comparison.} We report Pick-Score, CLIP-T, CLIP-F, WarpSSIM, and Q\textsubscript{edit} for 41-frame and 81-frame videos. Each entry is reported as $a / b$, corresponding to 41-frame and 81-frame results, respectively. $\uparrow$ indicates that higher is better.The ``-'' symbol indicates cases where a method cannot be evaluated at a given length due to memory limits or lack of support. We bold the \textbf{best value} for each metric and underline the \underline{second-best value}, both computed excluding FlowDirector (14B).
    }
    \vspace{-2mm}
    \resizebox{\textwidth}{!}{%
    \renewcommand{\arraystretch}{1.1}
    \setlength{\tabcolsep}{6pt}
    \begin{tabular}{cccccc}
        \toprule
        \multirow{2}{*}{\textbf{\small Method}} & 
        \multicolumn{3}{c}{\textbf{\textit{\small Perceptual Quality}}} & 
        \multicolumn{2}{c}{\textbf{\textit{\small Temporal Quality}}} \\
        \cmidrule(lr){2-4} \cmidrule(lr){5-6}
        & \textbf{\footnotesize Pick Score (\%) $\uparrow$}
        & \textbf{\footnotesize CLIP-T ($\times 10^{-2}$) $\uparrow$}
        & \textbf{\footnotesize CLIP-F ($\times 10^{-2}$) $\uparrow$}
        & \textbf{\footnotesize WarpSSIM ($\times 10^{-2}$) $\uparrow$}
        & \textbf{\footnotesize Q\textsubscript{edit} ($\times 10^{-6}$) $\uparrow$} \\
        \midrule
        \small FateZero~\cite{fatezero}
        & 20.41 / -
        & 32.01 / -
        & 92.25 / -
        & \underline{78.37} / -
        & 25.09 / - \\
        \small FLATTEN~\cite{flatten}
        & 20.84 / 20.18
        & \underline{33.56} / \underline{33.12}
        & 92.80 / 92.35
        & 77.44 / \underline{78.21}
        & \underline{26.01} / \underline{25.90} \\
        \small TokenFlow~\cite{tokenflow}
        & 20.99 / 20.63
        & 32.69 / 32.27
        & 93.82 / \underline{94.15}
        & 74.98 / 75.43
        & 24.51 / 24.34 \\
        \small RAVE~\cite{rave}
        & \underline{21.01} / \underline{20.76}
        & 33.25 / 33.10
        & 94.03 / 93.58
        & 76.32 / 75.91
        & 25.38 / 25.13 \\
        \small VideoDirector~\cite{videodirector}
        & 20.61 / -
        & 32.56 / -
        & \underline{95.48} / -
        & 75.89 / -
        & 24.70 / - \\
        \midrule
        \textbf{\small FlowDirector (1.3B)}
        & \textbf{21.82} / \textbf{21.69}
        & \textbf{34.64} / \textbf{34.63}
        & \textbf{97.34} / \textbf{96.90}
        & \textbf{78.49} / \textbf{79.10}
        & \textbf{27.19} / \textbf{27.31} \\
        \textbf{\small FlowDirector (14B)}
        & 22.61 / -
        & 34.95 / -
        & 97.30 / -
        & 79.86 / -
        & 28.67 / - \\
        \bottomrule
    \end{tabular}%
    }
    \label{tab:qc}
\vspace{-4mm}
\end{table*}

\section{Experiments}
\label{sec:experiments}

\subsection{Experimental Setups}

We build FlowDirector on the Wan-2.1 1.3B model~\cite{wan} and perform inversion-free video editing with 50 denoising steps without skip sampling. For Direction-Aware Flow Correction, we fix $\alpha = 0.25$ to amplify anti-parallel components and $\lambda = 0.25$ to soften the attention mask. Motion-Appearance Decoupling Correction uses task-dependent hyperparameters $(\zeta,\phi)$: for edits with strong motion we set $(\zeta,\phi)=(0.01,\,0.3)$, while for milder motion we use $(0.007,\,0.5)$. Differential Averaging Guidance (DAG) estimates a high-quality velocity from three stochastic samples, forms a baseline by averaging the two candidates with the lowest cosine similarity to this mean, \ie, $L_{HQ}=3, K=2$, and applies guidance with strength $w=2.75$, reusing a fixed noise realization in the last eight denoising steps.

For qualitative and quantitative evaluation, we construct 150 video-text editing pairs from Internet videos and the DAVIS~\cite{davis} dataset, covering insertion, deletion, and object editing. We evaluate at 41 and 81 frames and report results for both settings. Unless otherwise specified, all experiments for FlowDirector are run on a single NVIDIA H20 141G GPU. We compare against five state-of-the-art video editing methods: FateZero~\cite{fatezero}, FLATTEN~\cite{flatten}, TokenFlow~\cite{tokenflow}, RAVE~\cite{rave}, and VideoDirector~\cite{videodirector}, using their official implementations and default hyperparameters. All main results are obtained with the 1.3B model. Additional results for the 14B model and further implementation details are provided in the supplementary material.

\subsection{Qualitative Results}

We conduct a qualitative evaluation of FlowDirector on diverse video content under a variety of editing prompts. Our method handles large-scale object editing, video colorization, local attribute modification, object addition and removal, multi-object editing, and compositions of these operations (Figure~\ref{fig:teaser} and Figure~\ref{fig:qr}). For large-scale object editing, FlowDirector seamlessly converts the primary subject from one category to another while allowing substantial shape deformation (\eg, the first and second rows, brown bear'' $\rightarrow$ horse'', and the fifth and seventh rows, paper boat'' $\rightarrow$ duck''). For video colorization, it can colorize black-and-white videos while preserving the original structure and content (\eg, the third and fourth rows, colorful video''). For fine-grained object editing, FlowDirector modifies small objects while keeping the rest of the scene unchanged (\eg, watermelon'' $\rightarrow$ apple'' in the third and fourth rows). It also maintains complex motion, such as the continuous falling motion in the paper boat'' case, and in challenging long-occlusion scenarios it preserves the identity and appearance of the edited object consistently before and after the occlusion (\eg, the last two rows, bike'' $\rightarrow$ motorcycle'' with $\sim$20 occluded frames). Furthermore, FlowDirector adeptly handles attribute insertion on human subjects under strong appearance changes, simultaneously altering identity (Leonardo'' $\rightarrow$ Donald Trump''), gesture (raising the glass'' $\rightarrow$ giving a thumbs up''), and adding an extra red hat (fifth to seventh rows). Across these cases, the edits remain faithful to textual prompts, preserve details and motion in unedited regions, and exhibit strong temporal coherence and visual plausibility. We also compare FlowDirector with several state-of-the-art video editing methods (Figure~\ref{fig:compare}), where FlowDirector consistently produces higher-quality edits with fewer artifacts. More comprehensive qualitative results and detailed comparisons are provided in the supplementary material.

\subsection{Quantitative Results}

We conduct a quantitative comparison between FlowDirector and other state-of-the-art methods. Following prior work~\cite{fatezero, flatten, rave}, we evaluate text-content alignment using CLIP-T, the average CLIP~\cite{clip} embedding distance between the text prompt and video frames. Temporal coherence is measured by CLIP-F, the average pairwise frame CLIP similarity. Structure preservation during editing is assessed via WarpSSIM, the average SSIM between the final edited video and the source video warped using RAFT~\cite{raft} optical flow. Consistent with~\cite{flatten, rave}, we utilize the composite metric \textbf{${Q_{edit}}$} (WarpSSIM·CLIP-T) for a holistic evaluation of editing performance. Furthermore, we use Pick-Score~\cite{pick-score} to evaluate overall perceptual quality and prompt alignment based on human preferences.
Our experimental results (Table~\ref{tab:qc}) demonstrate that our proposed method significantly outperforms current state-of-the-art techniques regarding both text alignment and temporal consistency, and achieves strong performance on Pick-Score. Furthermore, our method achieves superior results on the comprehensive ${Q_{edit}}$ metric. This demonstrates that our method achieves the best overall results across diverse editing tasks.

\begin{table}[tbp]
    \centering
    \caption{Ablation results for Direction-Aware Flow Correction (DA-FC), Motion-Appearance Decoupling Correction(MAD-FC) and Differential Averaging Guidance (DAG). We highlight the \colorbox{myblue}{best} values for each metric.}
    \label{tab:ablation}
    \vspace{-2mm}
    \setlength{\tabcolsep}{4pt}
    \renewcommand{\arraystretch}{1}
    \small
    \begin{adjustbox}{max width=\columnwidth}
    \begin{tabular}{l c c c c}
    \toprule
    \textbf{Method}   & \textbf{CLIP-T $\uparrow$} & \textbf{CLIP-F $\uparrow$} & \textbf{WarpSSIM $\uparrow$} & \textbf{Q\textsubscript{edit} $\uparrow$} \\
    \midrule
    \small Director ODE (no skip)           
    & 34.59
    & 94.66
    & 62.71
    & 21.69 \\
    \small Director ODE (skip)           
    & 32.23 
    & 96.84 
    & \colorbox{myblue}{78.90}
    & 25.43 \\
    \small w/o DA-FC                
    & 32.25 
    & 95.70 
    & 78.83
    & 25.42 \\
    \small w/o MAD-FC               
    & \colorbox{myblue}{34.71}
    & 97.10 
    & 69.26 
    & 24.04 \\
    \small w/o DAG                  
    & 34.62 
    & 97.19
    & 78.32 
    & 27.11 \\
    \textbf{FlowDirector}    
    & 34.64
    & \colorbox{myblue}{97.34}
    & 78.49 
    & \colorbox{myblue}{27.19} \\
    \bottomrule
    \end{tabular}
    \end{adjustbox}
    \vspace{-2mm}
\end{table}

\begin{table}[tbp]
    \centering
    \caption{
        \textbf{Ablation results for Direction-Aware Flow Correction (DA-FC).}
        Para denotes the parallel component aligned with the source semantic direction. Therefore, w/o Para indicates removal of only this component, while a '-' value for $\alpha$ signifies no amplification of the opposite component. We highlight the \colorbox{myblue}{best} values for each metric. 
    }
    \vspace{-2mm}
    \label{tab:ablation-dafc}
    \setlength{\tabcolsep}{4pt}
    \renewcommand{\arraystretch}{1}
    \small
    \begin{adjustbox}{max width=\columnwidth}
    \begin{tabular}{l c c c c c}
    \toprule
    \textbf{Method} 
    & \textbf{$\alpha$}
    & \textbf{CLIP-T $\uparrow$} 
    & \textbf{CLIP-F $\uparrow$} 
    & \textbf{WarpSSIM $\uparrow$} 
    & \textbf{Q\textsubscript{edit} $\uparrow$} \\
    \midrule
    \small w/o DA-FC           
    & --
    & 32.25 
    & 95.70 
    & 78.83
    & 25.42 \\
    \small + w/o Para                
    & --
    & 33.02 
    & 96.51 
    & \colorbox{myblue}{78.85}
    & 26.04 \\
    \small + w/o Para          
    & 0.05
    & 33.14
    & 96.49 
    & 78.80 
    & 26.11 \\
    \small + w/o Para               
    & 0.15
    & 34.57 
    & 97.30
    & 78.48 
    & 27.13 \\
    \small \textbf{+ w/o Para}   
    & \textbf{0.25}
    & \colorbox{myblue}{34.64}
    & \colorbox{myblue}{97.34}
    & 78.49 
    & \colorbox{myblue}{27.19} \\
    \bottomrule
    \end{tabular}
    \end{adjustbox}
    \vspace{-4mm}
\end{table}

\subsection{Ablation Study}
\label{sec:ablation_study}

\textbf{Direction-Aware Flow Correction.} As shown in Table~\ref{tab:ablation} and Table~\ref{tab:ablation-dafc}, removing the parallel component that is aligned with the source direction effectively improves the consistency of the editing results. As the scaling factor $\alpha$ increases, the semantic change of the edited object becomes progressively stronger. However, the optical-flow-based metric keeps decreasing. We attribute this to the fact that stronger edits introduce pronounced shape deformations, so warping the edited frame using optical flow estimated from the source video leads to severe misalignment, which in turn lowers the WarpSSIM score.
In addition, we performed a qualitative ablation analysis of DA-FC and present an ablation study on mask generation in the supplementary materials.

\begin{figure}[tb]
    \centering
    \includegraphics[width=0.88\linewidth]{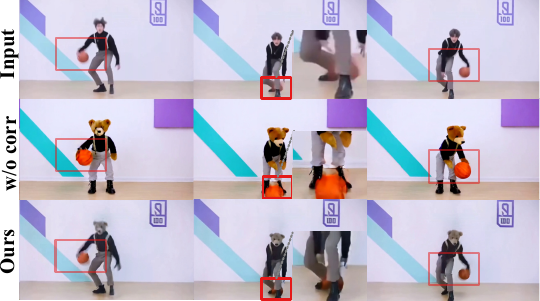}
    \vspace{-2mm}
    \caption{\textbf{Ablation study of Motion-Appearance Decoupling Correction.} Without MAD-FC, the edited video exhibited severe distortion. After using MAD-FC correction, the motion of the edited video remained largely consistent with the original video.}
    \label{fig:ma-corr}
    \vspace{-2mm}
\end{figure}

\noindent \textbf{Motion-Appearance Decoupling Correction.} Figure~\ref{fig:ma-corr} illustrates that FlowDirector (Direct ODE) suffers from motion distortion and drift in complex action editing due to accumulated inter-frame errors (red boxes, w/o corr row: the ball incorrectly appears in front), whereas MAD-FC preserves coherent motion even under fast movement with occlusion (red boxes, ours row: the crossover dribble is correctly maintained). Quantitative results in Table~\ref{tab:ablation} show a clear improvement in WarpSSIM computed from the source-video optical flow, confirming that MAD-FC better preserves motion consistency before and after editing. In the supplementary material, we further provide ablations on the MAD-FC parameters $\zeta$ (overall guidance strength) and $\phi$ (appearance alignment strength), offering a detailed analysis of their impact on the final editing quality.

\begin{figure}[tb]
    \centering
    \includegraphics[width=0.88\linewidth]{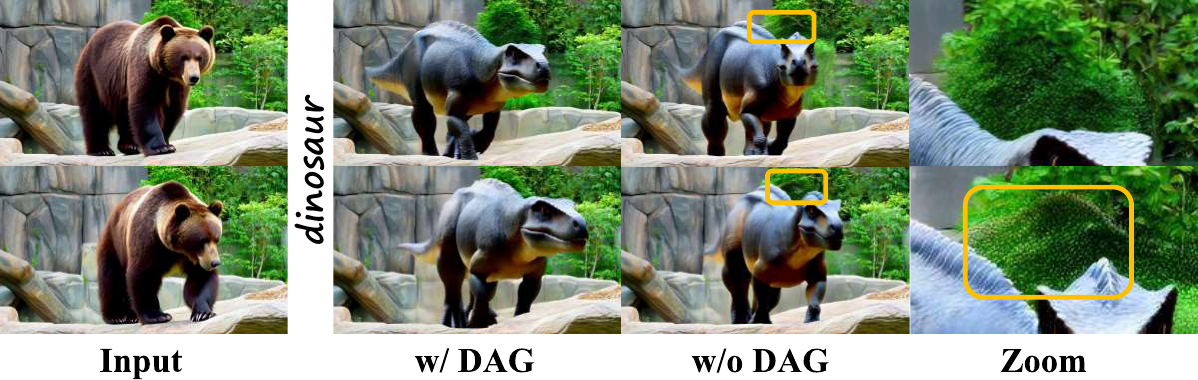}
    \vspace{-2mm}
    \caption{\textbf{Ablation study of Differential Averaging Guidance.} In the ``Zoom'' column: Top row is with DAG, bottom row is without DAG.}
    \label{fig:dag}
    \vspace{-5mm}

\end{figure}

\noindent \textbf{Differential Averaging Guidance.} As shown in Figure~\ref{fig:dag}, without DAG the high variance of sampling noise makes the editing trajectory jitter across timesteps, which introduces visible artifacts and texture flicker. With DAG, these issues are largely suppressed, and the textures remain temporally stable and consistent with the underlying motion. As reported in Table~\ref{tab:ablation}, enabling DAG leads to slight improvements on all metrics. The numerical gains are relatively modest because many perceptual degradations, such as artifacts, texture flicker, and other local inconsistencies, are not fully reflected by these metrics. For example, optical-flow-based WarpSSIM and CLIP-T are often insensitive even when such issues are clearly noticeable to human observers. In the supplementary materials, we provide a detailed analysis of runtime and resource consumption, and propose several optimization and acceleration strategies for DAG.

\section{Conclusion}

In this paper, we propose FlowDirector, a novel training-free video editing framework that operates without inversion, instead modeling video transformation as a continuous evolution in data space. We propose three training-free flow correction strategies to optimize the editing path. Direction-aware flow correction amplifies components opposite to the source semantics to promote thorough editing. Motion-appearance decoupling flow correction models the motion consistency before and after editing as an optimizable energy term and continuously minimizes it to preserve temporal structure. Differential averaging guidance achieves variance reduction close to that of multiple averaging with fewer averaging operations and improves trajectory stability. Extensive experiments show that FlowDirector achieves state-of-the-art performance in text-driven video editing.

{
    \small
    \bibliographystyle{ieeenat_fullname}
    \bibliography{main}
}

\clearpage
\setcounter{page}{1}
\maketitlesupplementary
\appendix

This supplementary material provides comprehensive implementation details, in-depth algorithmic descriptions, and extensive experimental analyses to further validate the effectiveness of FlowDirector. We begin by outlining the specific experimental configurations and hyperparameter settings in \Cref{sec:appendix_a}, followed by the complete inference pseudocode in \Cref{sec:appendix_algo}. \Cref{sec:mask_generation} details the process of mask generation via cross-attention maps. Subsequently, we present a deeper investigation into our core contributions, including a qualitative ablation analysis of the Direction-Aware Flow Correction in \Cref{sec:appendix_dafc}, a parameter study of the Motion-Appearance Decoupling Flow Correction in \Cref{sec:appendix_madfc}, and a detailed efficiency analysis of the Differential Averaging Guidance strategy in \Cref{sec:appendix_dag}. Finally, we discuss current limitations in \Cref{sec:appendix_limitation} and showcase an extensive gallery of additional qualitative results across diverse editing scenarios in \Cref{sec:more_results}.

\section{Detailed Experimental Settings}
\label{sec:appendix_a}

Our implementation leverages the pre-trained Wan-2.1 1.3B model~\cite{wan} as the foundational backbone. The editing procedure is executed over a full 50-step denoising trajectory without employing skip sampling strategies. During inference, we disable Classifier-Free Guidance (CFG) for the source video branch, whereas the target video generation utilizes a fixed CFG scale of 10.5. To better align the temporal dynamics, a timestep shift of 12 is applied throughout the sampling process. Regarding the specific hyperparameters of FlowDirector, the Direction-Aware Flow Correction is configured with an amplification factor $\alpha = 0.25$ and a softening coefficient $\lambda = 0.25$. For mask generation, we apply average pooling with a kernel size of 9 to ensure boundary smoothness. The Motion-Appearance Decoupling Correction adapts its regularization strength $(\zeta,\phi)$ according to the editing magnitude: we assign $(\zeta,\phi)=(0.01,\,0.3)$ for edits involving significant motion changes, and $(0.007,\,0.5)$ for those with subtler dynamics. Furthermore, the Differential Averaging Guidance (DAG) is computed using a guidance weight of $\omega=2.75$, deriving a robust velocity estimate from $L_{HQ}=3$ stochastic samples and establishing the baseline from the $K=2$ candidates with the lowest cosine similarity. To stabilize the final output details, the noise realization remains frozen during the last eight denoising steps.

\section{FlowDirector Inference Algorithm}
\label{sec:appendix_algo} 

We present the complete inference procedure of FlowDirector in Algorithm \ref{alg:flowdirector}. Our framework operates in an inversion-free manner, initializing the denoising trajectory directly from the source video $X^{\text{src}}$. At each timestep $t$, the update process integrates three core strategies: 
(1) \textbf{Direction-Aware Flow Correction (DA-FC)} is applied during the candidate generation phase, where we amplify anti-parallel flow components to facilitate structural changes; 
(2) \textbf{Differential Averaging Guidance (DAG)} aggregates these corrected candidates to estimate a robust editing velocity $v_{\text{edit}}$, utilizing a baseline formed by high-variance samples to reduce trajectory jitter; 
and (3) \textbf{Motion-Appearance Decoupling Flow Correction (MAD-FC)} rectifies the state update. Specifically, we estimate the clean data states $\hat{x}_0$ and their temporal averages (anchors $A_h$) to enforce motion consistency while allowing appearance changes via the parameter $\phi$. The final state $Z_{i-1}$ is updated by combining the masked editing flow with this decoupling correction term.

\begin{algorithm}[t]
\caption{FlowDirector Inference Algorithm}
\label{alg:flowdirector}
\small
\renewcommand{\baselinestretch}{1.15}\selectfont 

\textbf{Input:} Source $X^{\text{src}}$, Prompts $c_{\text{src}}, c_{\text{tar}}$, Steps $N$, Hyperparams $\alpha, \zeta, \omega, L_{\text{HQ}}, K$. \\
\textbf{Output:} Edited Video $Z_0$.

\begin{algorithmic}[1]
\STATE $Z_N \leftarrow X^{\text{src}}$
\FOR{$i = N$ \textbf{to} $1$}
    \STATE $t \leftarrow t_i, \quad \Delta t \leftarrow t_{i-1} - t_i$
    \STATE $\mathcal{V} \leftarrow \emptyset$
    
    \FOR{$k = 1$ \textbf{to} $L_{\text{HQ}}$}
        \STATE Sample $\epsilon_k \sim \mathcal{N}(0, I)$ to obtain $z_t^{\text{src}}, z_t^{\text{tar}}$
        \STATE $v_{\text{src}} \leftarrow v_\theta(z_t^{\text{src}}, t, c_{\text{src}}), \quad v_{\text{tar}} \leftarrow v_\theta(z_t^{\text{tar}}, t, c_{\text{tar}})$
        \STATE $v_{\text{raw}} \leftarrow v_{\text{tar}} - v_{\text{src}}$
        \STATE $v_{\parallel} \leftarrow \frac{\langle v_{\text{raw}}, v_{\text{src}} \rangle}{\|v_{\text{src}}\|^2} v_{\text{src}}, \quad v_{\perp} \leftarrow v_{\text{raw}} - v_{\parallel}$
        
        \STATE $M_{\text{opp}} \leftarrow \mathbb{1}(\langle v_{\text{raw}}, v_{\text{src}} \rangle < 0)$ \hfill // Element-wise indicator
        
        \STATE $\tilde{v}_k \leftarrow v_{\perp} + M_{\text{opp}} \odot (1 + \alpha) v_{\parallel}$
        \STATE $\mathcal{V} \leftarrow \mathcal{V} \cup \{ \tilde{v}_k \}$
    \ENDFOR

    \vspace{0.5em}
    \STATE $v_{\text{HQ}} \leftarrow \frac{1}{L_{\text{HQ}}} \sum_{\tilde{v} \in \mathcal{V}} \tilde{v}$
    \vspace{0.3em}
    \STATE $S_k \leftarrow \frac{\langle \tilde{v}_k, v_{\text{HQ}} \rangle}{\|\tilde{v}_k\| \|v_{\text{HQ}}\|} \quad \forall \tilde{v}_k \in \mathcal{V}$
    \vspace{0.3em} 
    \STATE $\mathcal{I}_{K} \leftarrow \text{Indices of } K \text{ smallest values in } S$
    \STATE $v_{\text{BL}} \leftarrow \frac{1}{K} \sum_{k \in \mathcal{I}_{K}} \tilde{v}_k$
    \STATE $v_{\text{edit}} \leftarrow v_{\text{HQ}} + \omega \cdot (v_{\text{HQ}} - v_{\text{BL}})$

    \vspace{0.5em}
    \STATE $\hat{x}_0^{\text{src}} \leftarrow Z_i - t \cdot v_\theta(Z_i, t, c_{\text{src}})$ \hfill // (C, T, H, W)
    \STATE $\hat{x}_0^{\text{tar}} \leftarrow Z_i - t \cdot v_\theta(Z_i, t, c_{\text{tar}})$ \hfill // (C, T, H, W)
    \STATE $A_h^{\text{src}} \leftarrow \text{Mean}_T(\hat{x}_0^{\text{src}}), \quad A_h^{\text{tar}} \leftarrow \text{Mean}_T(\hat{x}_0^{\text{tar}})$
    
    \vspace{0.5em}
    \STATE $\hat{v}_{\text{final}} \leftarrow v_{\text{edit}} \odot \text{Mask}(c_{\text{src}}, c_{\text{tar}})$
    \STATE $Z_{i-1} \leftarrow Z_i + \Delta t \cdot \hat{v}_{\text{final}} - \zeta \Big[\hat{x}_0^{\text{tar}} -\hat{x}_0^{\text{src}} - \phi (A_h^{\text{src}} - A_h^{\text{tar}})\Big]$

\ENDFOR
\RETURN $Z_0$
\end{algorithmic}
\end{algorithm}

\begin{figure}[t]
    \centering
    \includegraphics[width=\linewidth]{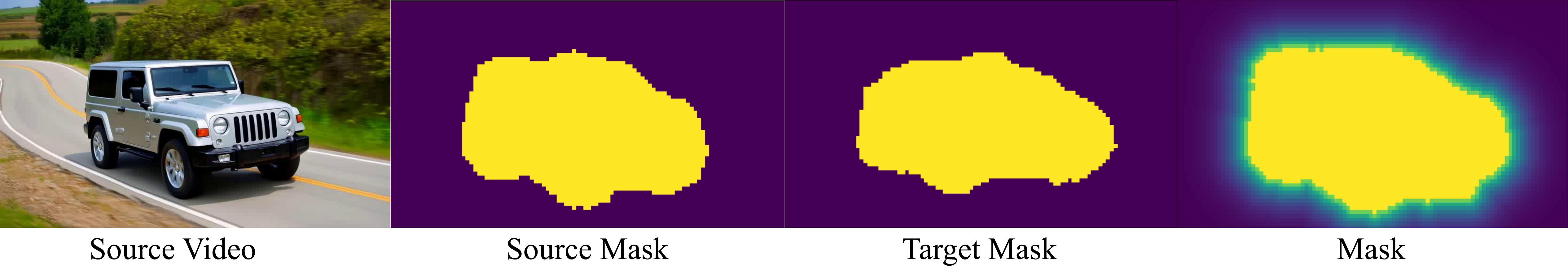}
    \caption{The final mask is obtained by merging the Source Mask and the Target Mask and softening the edges. Lighter colors indicate lower attention values.}
    \label{fig:spfa_ab}
\end{figure}

\section{Mask Generation via Cross-Attention Maps}
\label{sec:mask_generation}

\begin{figure*}[tb]
  \centering
  \includegraphics[width=\textwidth]{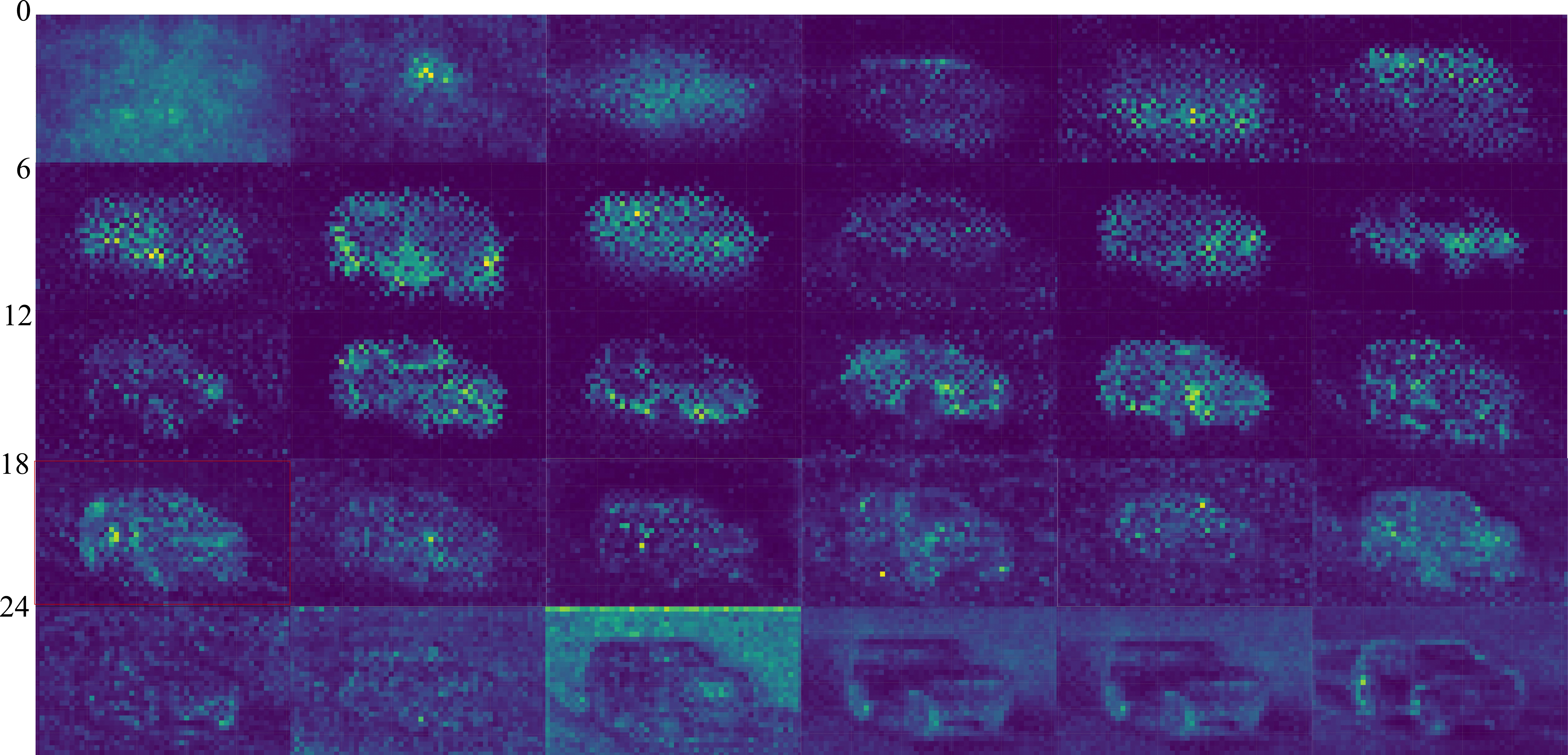}
  \caption{Visualization of the cross-attention maps of the keyword ``jeep'' in the source prompt across different DiT blocks. The attention map of the 18th block clearly outlines the shape of the jeep.}
  \label{fig:src}
\end{figure*}

\begin{figure*}[tb]
  \centering
  \includegraphics[width=\textwidth]{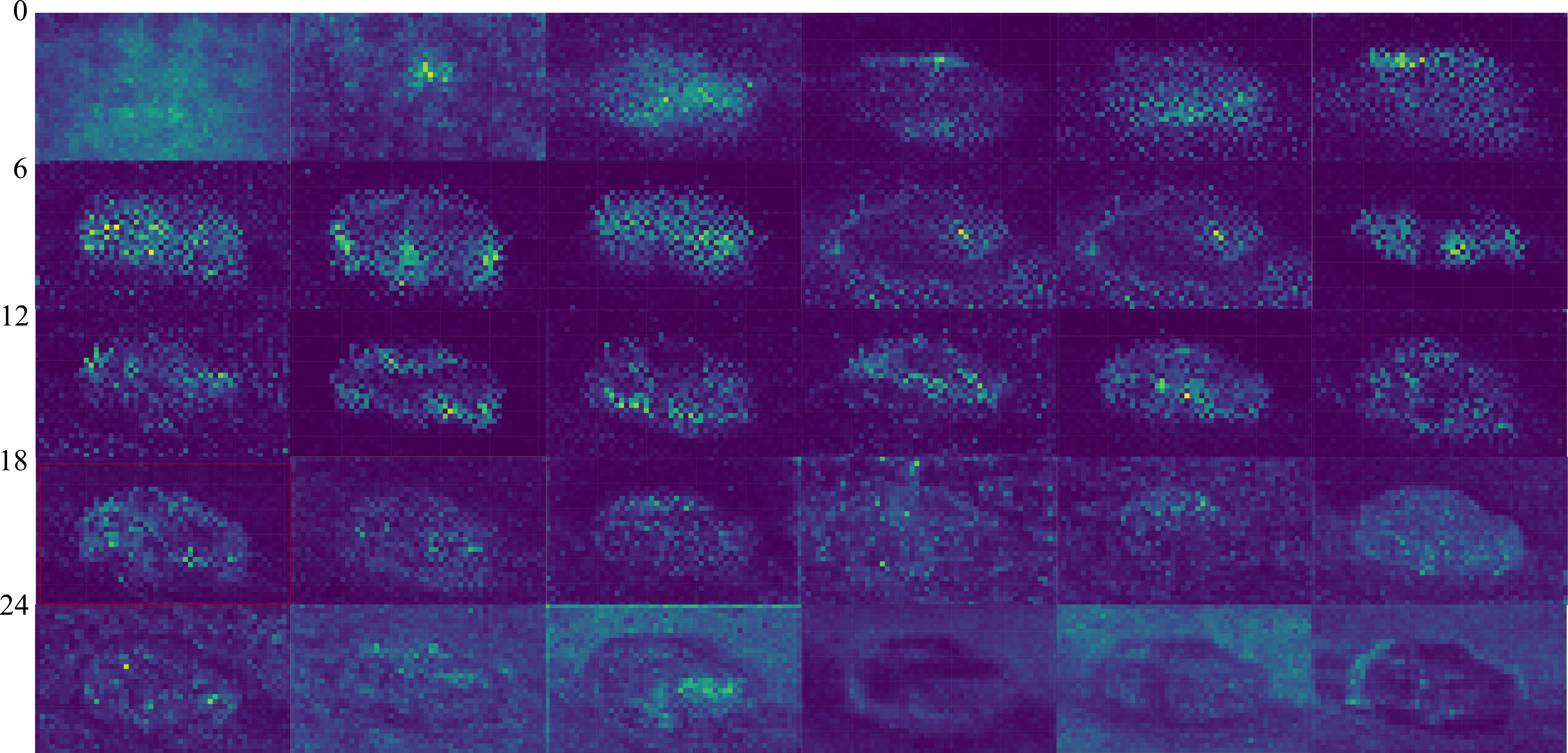}
  \caption{Visualization of the cross-attention maps of the keyword ``Porsche car'' in the source prompt across different DiT blocks. The attention map of the 18th block clearly outlines the car.}
  \label{fig:tar}
\end{figure*}

Although the editing flow $V_{\text{edit}}$ effectively drives the semantic transformation, applying it globally can cause unintended modifications in the background. To address this, we construct an explicit spatial mask by leveraging the intrinsic localization capabilities of Diffusion Transformers (DiTs). By visualizing the cross-attention maps across different network depths (see Figure~\ref{fig:src} and Figure~\ref{fig:tar}), we empirically identified the \textbf{18th block} as the optimal block. The maps extracted from this block exhibit high activation concentrations on object structures, enabling us to cleanly separate editable regions from the background.

\noindent \textbf{Attention Extraction and Aggregation.} 
During the denoising step $t$, we perform a forward pass with the source prompt $c_{\text{src}}$ and extract the cross-attention map $\mathbf{A} \in \mathbb{R}^{B \times N_h \times L_{\text{vis}} \times L_{\text{text}}}$. Here, $N_h$ denotes the number of attention heads, while $L_{\text{vis}}$ and $L_{\text{text}}$ represent the visual and textual token counts. Given a set of indices $S$ corresponding to the key editing tokens in the prompt (e.g., ``jeep''), we aggregate the attention scores to obtain a consolidated spatial map $\mathbf{a} \in \mathbb{R}^{L_{\text{vis}}}$:
\begin{equation}
\mathbf{a}_i = \frac{1}{N_h|S|} \sum_{h=1}^{N_h} \sum_{j \in S} \mathbf{A}^{(h)}_{i,j}.
\label{eq:agg}
\end{equation}
This vector $\mathbf{a}$ is then reshaped into a spatiotemporal patch grid $\mathcal{G} \in \mathbb{R}^{F_p \times H_p \times W_p}$.

\noindent \textbf{Mask Construction and Refinement.} 
To generate a robust binary mask $\mathbf{M}_{\text{src}}$, we process $\mathcal{G}$ through the following steps:
\begin{enumerate}
    \item \textbf{Spatial Smoothing:} We apply 2D average pooling with a \textbf{kernel size of 9} to each frame in $\mathcal{G}$. This mitigates high-frequency noise in the raw attention maps.
    \item \textbf{Upsampling:} The smoothed map is upsampled to the original video resolution via trilinear interpolation and replicated along the channel dimension to form $\mathcal{A} \in \mathbb{R}^{C \times T \times H \times W}$.
    \item \textbf{Binarization:} We perform a numerically stable normalization $\hat{\mathcal{A}} = \mathcal{A} / (\max(\mathcal{A}) + \varepsilon)$ and derive a binary mask using an adaptive threshold $\tau = \mathrm{mean}(\hat{\mathcal{A}})$:
    \begin{equation}
    (\mathbf{M}_{\text{src}})_{c,t}(x,y)=
    \begin{cases}
    1, & \text{if } \hat{\mathcal{A}}_{c,t}(x,y)\ge \tau,\\
    0, & \text{otherwise.}
    \end{cases}
    \end{equation}
\end{enumerate}
An analogous procedure yields the target mask $\mathbf{M}_{\text{tar}}$ from the target prompt $c_{\text{tar}}$. The comprehensive editing region is defined as the union $\mathbf{M} := \mathbf{M}_{\text{src}} \cup \mathbf{M}_{\text{tar}}$.

\noindent \textbf{Soft Blending.} 
To ensure a seamless transition between the edited object and the background, we soften the binary mask using a Euclidean distance transform $d_{c,t}(x,y)$ computed on the background region. The final soft mask $\widetilde{\mathbf{M}}$ is formulated as:
\begin{equation}
\widetilde{\mathbf{M}}_{c,t}(x,y) = \mathbf{M}_{c,t}(x,y) + \left(1 - \mathbf{M}_{c,t}(x,y)\right) e^{-\lambda \, d_{c,t}(x,y)}.
\end{equation}
We set the decay rate $\lambda = 0.25$. This mask directly modulates the editing velocity field, freezing irrelevant regions while preserving the structural integrity of the edited subject:
\begin{equation}
\tilde{V}_{\text{edit}} = V_{\text{edit}} \odot \widetilde{\mathbf{M}}.
\end{equation}

\begin{figure}[tb]
    \centering
    \includegraphics[width=0.95\linewidth]{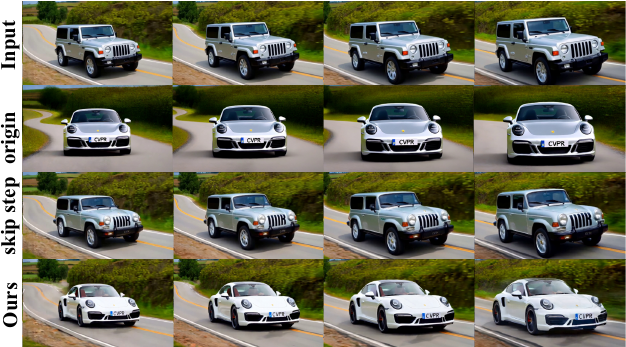}
    \caption{\textbf{Ablation study of Direction-Aware Flow Correction.} Without DA-FC, it is difficult to achieve an effective balance between editing strength and consistency. In contrast, incorporating DA-FC enables significant semantic modifications of the target object while effectively preserving irrelevant regions and maintaining motion consistency. Our modules are crucial for achieving high-quality video editing.}
    \label{fig:da-corr}
\end{figure}

\section{Qualitative Ablation Analysis Of Direction-Aware Flow Correction}
\label{sec:appendix_dafc}

We conduct a qualitative analysis to investigate the impact of Direction-Aware Flow Correction (DA-FC) on editing fidelity and structural integrity. As illustrated in Figure~\ref{fig:da-corr}, relying solely on the basic FlowDirector (Direct ODE) presents a dilemma. When configured without skip steps, the direct integration accumulates errors along the path, leading to severe deviations in both appearance and motion consistency compared to the source video. Conversely, employing skip steps to mitigate this drift imposes excessive constraints on the generative trajectory, which heavily restricts the editing magnitude and fails to produce significant structural changes.

Consequently, without DA-FC, the editing process struggles to strike an effective balance between semantic transformation and content preservation. Our Direction-Aware Flow Correction resolves this by intervening at the velocity level: it amplifies the anti-parallel components essential for structural editing while suppressing the parallel components that contribute to drift. This enables FlowDirector to achieve robust semantic transformations without the consistency degradation seen in the full ODE or the conservative limitations of skip sampling.

\section{Ablation Study of Motion-Appearance Decoupling Flow Correction}
\label{sec:appendix_madfc}

The Motion-Appearance Decoupling Flow Correction (MAD-FC) module serves a critical role in balancing source motion fidelity with target appearance transformation. This balance is governed by two key hyperparameters: the appearance anchor coefficient $\phi$ and the overall correction strength $\zeta$. In this section, we analyze their individual impacts on editing quality based on our empirical observations.

\begin{figure}[tb]
    \centering
    \includegraphics[width=\linewidth]{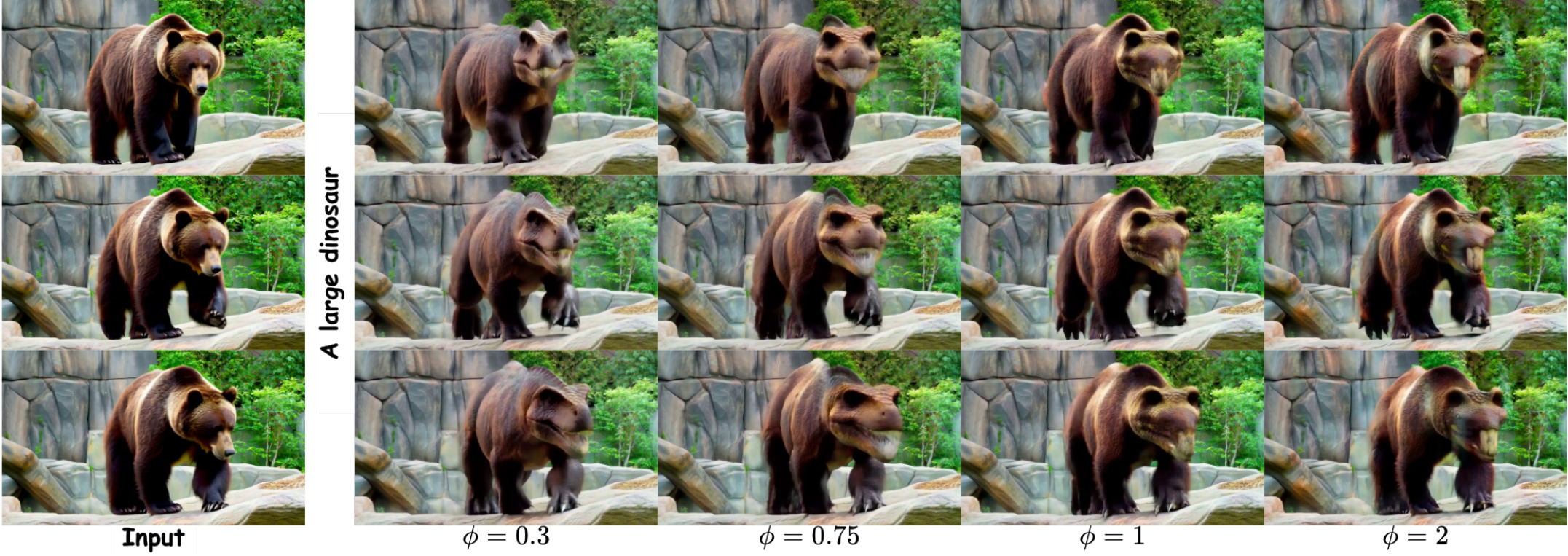}
    \caption{\textbf{Ablation study on the appearance anchor coefficient $\phi$.} We fix $\zeta=0.003$ and vary $\phi$ in the ``bear $\to$ dinosaur'' task. A very high $\phi$ (2.0) imposes excessive source constraints, causing the result to revert to the original bear appearance. As $\phi$ decreases, the constraints relax, allowing the dinosaur features to emerge. $\phi=0.3$ successfully achieves the semantic transformation while maintaining the original motion pattern.}
    \label{fig:ablation_phi}
\end{figure}

We first examine the influence of the appearance anchor coefficient $\phi$. This parameter regulates the adherence to the source video's visual attributes. As illustrated in Figure~\ref{fig:ablation_phi}, setting $\phi$ to an excessively high value (\eg, $\phi=2$) imposes rigid constraints derived from the source appearance anchors. Consequently, the generated output remains visually nearly identical to the original bear, effectively suppressing the desired semantic transformation. Even at $\phi=1$, the result retains significant bear-like morphological features, such as the head shape and fur texture. Conversely, lowering $\phi$ relaxes these constraints, allowing the target semantics to manifest. We observe that $\phi=0.3$ strikes an optimal balance, facilitating a successful morphological transformation into a dinosaur—characterized by changes in skin texture and body structure—while preserving the underlying walking motion of the original video.

\begin{figure}[tb]
    \centering
    \includegraphics[width=\linewidth]{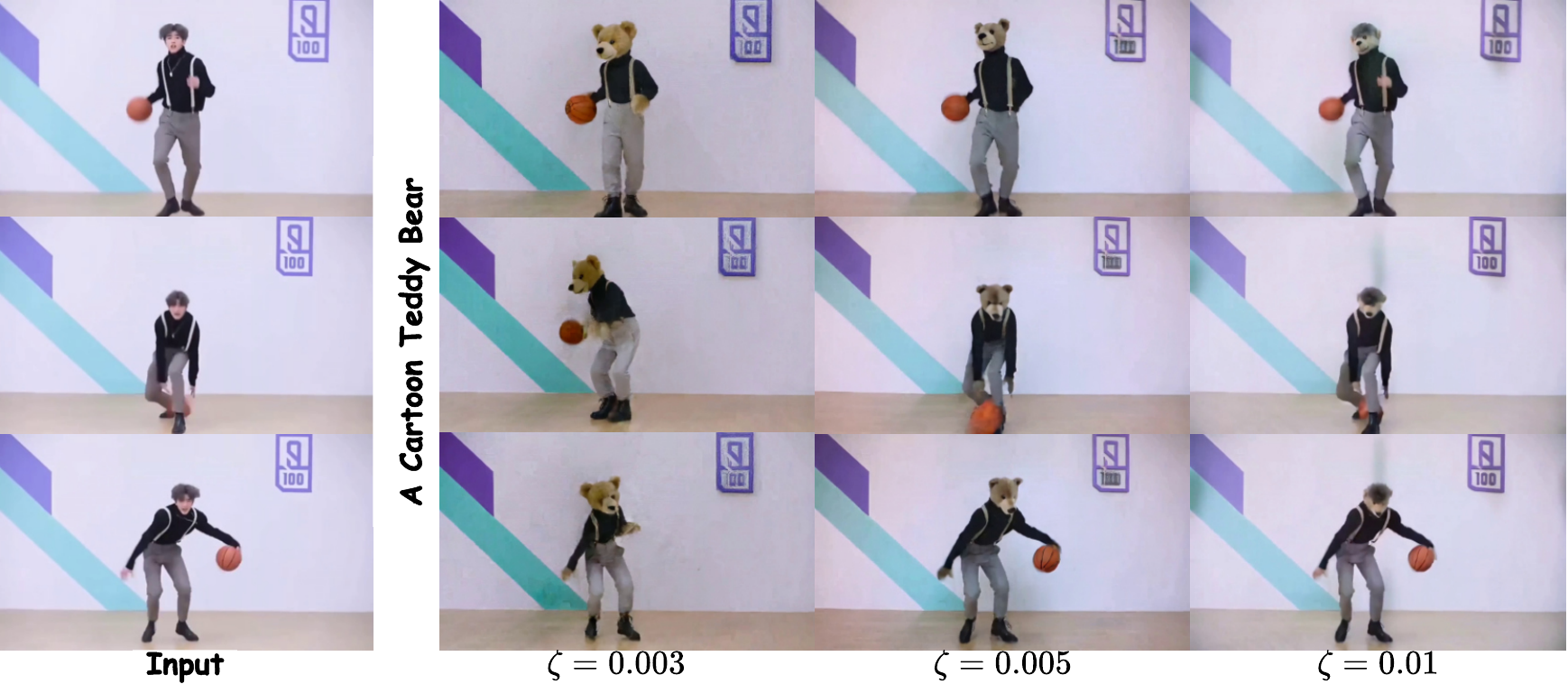}
    \caption{\textbf{Ablation study on the correction strength $\zeta$.} We fix $\phi=0.3$ and vary $\zeta$. Increasing $\zeta$ from 0.003 to 0.01 significantly improves motion alignment with the source input, ensuring that the edited character strictly follows the original movements.}
    \label{fig:ablation_zeta}
\end{figure}

Next, we investigate the correction strength $\zeta$, which determines the intensity of the motion consistency enforcement. As shown in Figure~\ref{fig:ablation_zeta}, a low correction strength (\eg, $\zeta=0.003$) provides insufficient guidance to counteract the stochastic variance of the diffusion process. This results in slight pose misalignments and temporal instability, where the edited character fails to strictly follow the source motion. Increasing $\zeta$ to $0.01$ significantly improves alignment, ensuring the edited subject's posture and trajectory align precisely with the source video. Based on these findings, we utilize a configuration of $(\zeta=0.01, \phi=0.3)$ for edits involving complex structural changes, and adjust to $(0.007, 0.5)$ for milder motion scenarios to prioritize stability.

\begin{figure*}[ht]
  \centering
  \includegraphics[width=0.95\textwidth]{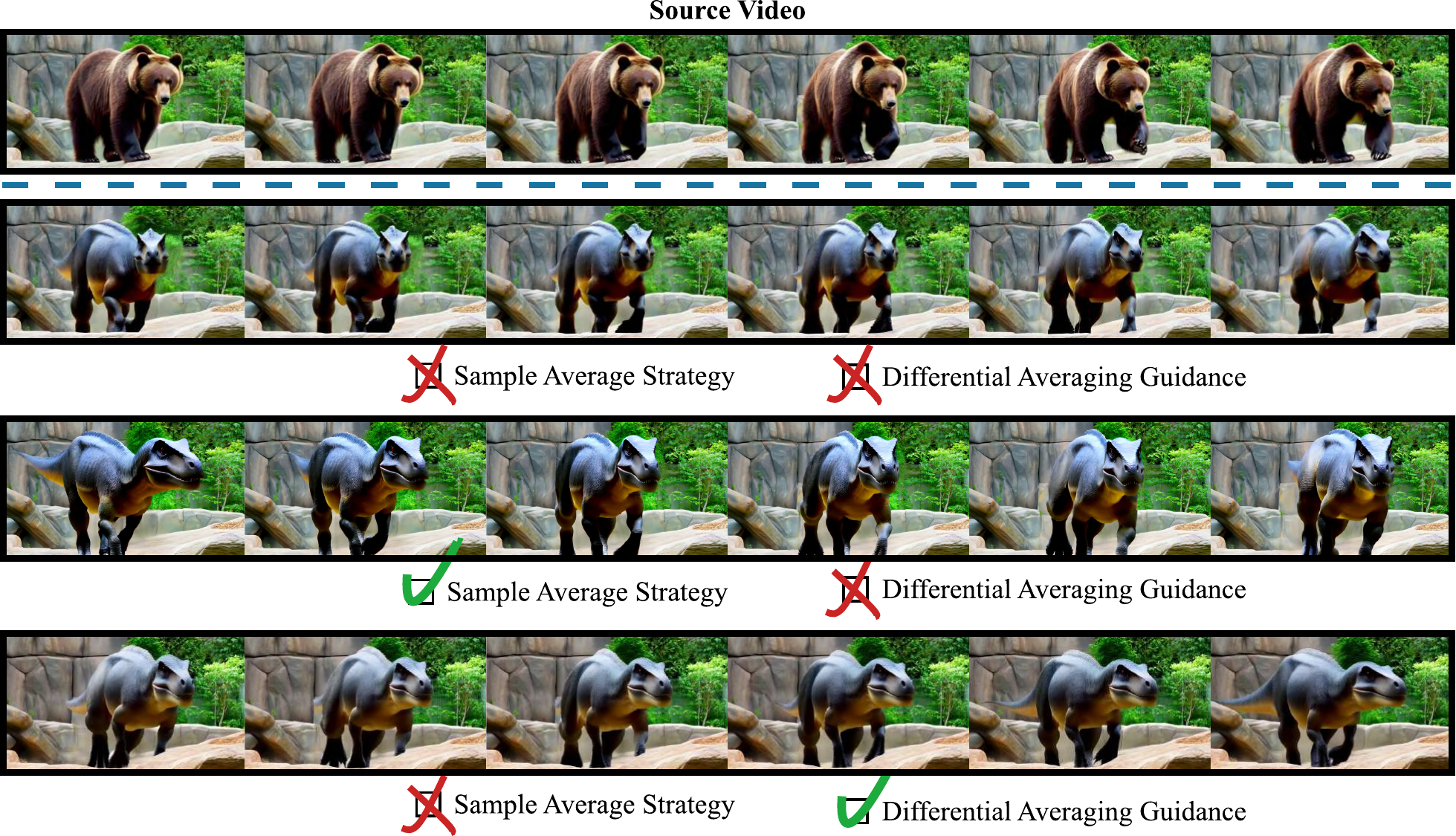}
  \caption{\textbf{Qualitative comparison between the editing results of a multi-round inference averaging strategy and using a DAG.} The Sample Average strategy is set to use a regular averaging strategy for 20 rounds of iterative inference at every denoising step to obtain the editing flow. The DAG setting uses 4 rounds of iterative inference to obtain a high-quality estimate and perform reinforcement-guided generation of the editing flow. Best viewed zoomed in.}
  \label{fig:abdag}
\end{figure*}

\begin{figure}[ht]
    \centering
    \includegraphics[width=\linewidth]{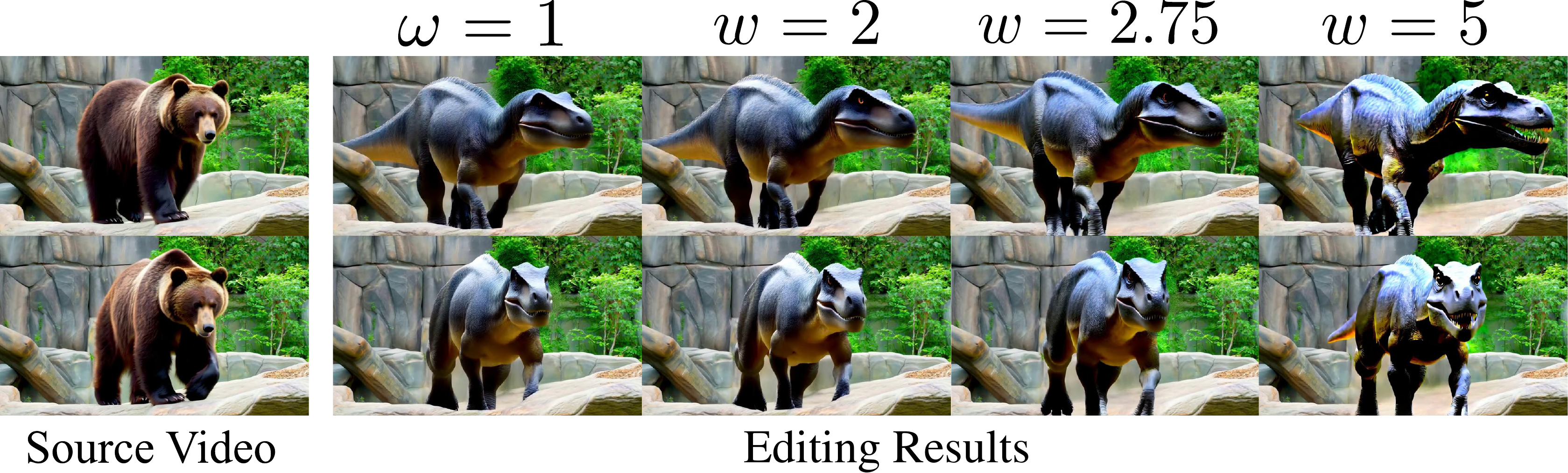}
    \caption{\textbf{Ablation study of guidance strength $\omega$.} The $\omega$ controls the guidance strength of the differential signal. By enhancing the differential signal, artifacts can be effectively eliminated and the editing results can be optimized. We use $\omega = 2.75$ as the default value. Best viewed zoomed in.}
    \label{fig:omega}
\end{figure}

\begin{figure}
    \centering
    \includegraphics[width=\linewidth]{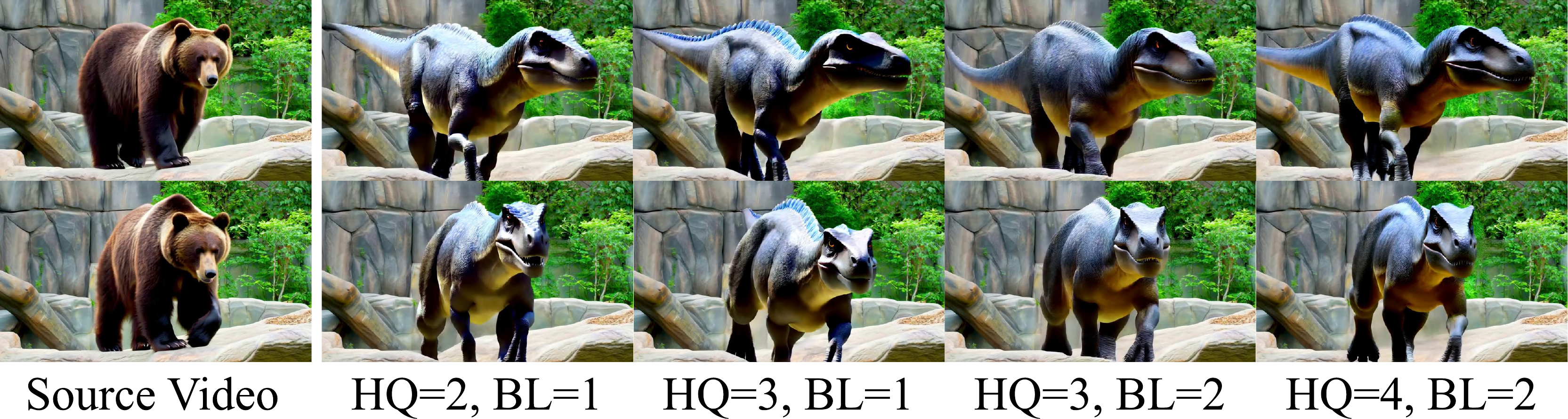}
    \caption{By taking different averages to construct high-quality estimates and baseline estimates, different guidance enhancement effects are produced. Best viewed zoomed in.}
    \label{fig:hqbl}
\end{figure}

\begin{figure*}[tb]
  \centering
  \includegraphics[width=0.95\textwidth]{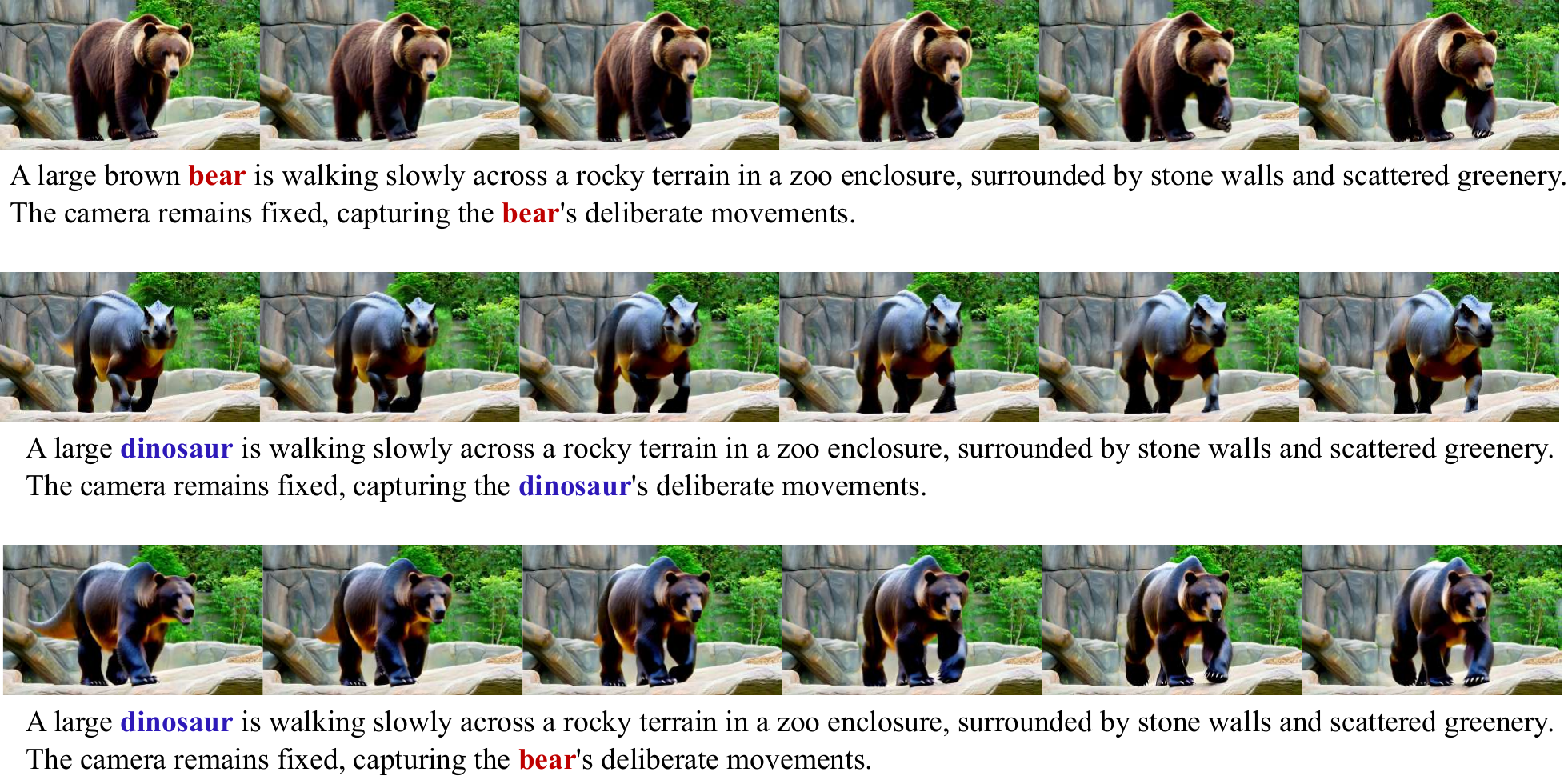}
  \caption{\textbf{An example of editing failure due to incomplete target text replacement.} When attempting to edit a ``bear'' into a ``dinosaur,'' if the target prompt erroneously retains descriptions of the ``bear'' (\eg, ``...capturing the bear's deliberate movements'' instead of a full replacement with dinosaur-related descriptions), the edited video exhibits significant residual features of the original ``bear.''}
  \label{fig:fail_text}
\end{figure*}

\section{Detailed Analysis of Differential Averaging Guidance}
\label{sec:appendix_dag}

A conventional averaging method involves using multiple rounds of iterative inference to obtain several flows, each indicating different editing directions. These flows are then averaged to derive a more robust, consolidated direction, which is subsequently used to update our video features. In this section, we compare our Differential Averaging Guidance (DAG) approach with the aforementioned conventional averaging strategy to demonstrate the effectiveness and efficiency of our DAG method. As illustrated in Figure~\ref{fig:abdag}, when editing the source video (\eg, a bear) into the target video (\eg, a dinosaur), the regions experiencing significant editing changes are primarily concentrated on the bear's back. When no specific strategy is employed, noticeable artifacts and inter-frame texture flickering appear in the area corresponding to the original bear's back (\eg, the second row, Figure~\ref{fig:abdag}).

\begin{table}[tb]
\centering
\small
\caption{\textbf{Efficiency Comparison.} We report the inference time and peak GPU memory usage for editing a 41-frame video on a single NVIDIA H800 80G GPU. The upper section compares existing SOTA methods, while the lower section analyzes the efficiency of different strategies within our framework. The symbol ``--'' indicates cases where the method exceeded the single-GPU memory limit and required specific optimization strategies to execute; consequently, these metrics are omitted to ensure a fair comparison of native performance.}
\label{tab:efficiency_ablation}
\renewcommand{\arraystretch}{1.15} 
\begin{tabular}{l c c} 
\toprule
\textbf{Method} & \textbf{Editing Time} & \textbf{GPU Memory} \\
\midrule
FateZero~\cite{fatezero} & -- & -- \\
FLATTEN~\cite{flatten} & 6 min 13s & 44.7GB \\
TokenFlow~\cite{tokenflow} & 1 min 15s & 29.7GB \\
RAVE~\cite{rave} & 5min 46s & 26.4GB \\
VideoDirector~\cite{videodirector} & -- & -- \\
\midrule
w/o DAG & 57s  & 18 GB \\
Conventional Averaging & 19 min 3s & $\approx$18 GB \\
\textbf{w/ DAG} & 2 min 54s & $\approx$18 GB \\
\bottomrule
\end{tabular}
\end{table}

When the conventional averaging strategy is applied, using the results from twenty rounds of iterative inference (Figure~\ref{fig:abdag}, third column), these artifacts are largely eliminated. However, this comes at a significant computational cost ~\textbf{(editing a 41-frame video requires about 19 minutes on a single NVIDIA H800 80G GPU).}.

In contrast, our DAG approach requires only four rounds of iterative inference to achieve high-quality estimation ~\textbf{(editing a 41-frame video requires about 3 minutes on a single NVIDIA H800 80G GPU)}. This estimation is used to generate a differential signal that guides and reinforces our editing flow, producing comparable or even superior results. As shown in Figure~\ref{fig:abdag} (fourth row), compared to the twenty-round averaging result, our method more effectively eliminates artifacts, resolves issues of incomplete editing, and the resulting dinosaur exhibits a morphology more distinct from the original bear. This provides strong evidence for the efficiency and effectiveness of our proposed method.

\paragraph{Effect of the Guidance Strength $\omega$.}
To determine the optimal guidance strength $\omega$, we conducted a comprehensive ablation study. The primary objective of this analysis was to investigate the impact of varying $\omega$ values on the quality of the video editing results. As illustrated in Figure~\ref{fig:omega}, our experiments reveal a clear correlation between the guidance strength and the final output.
Specifically, as the value of $\omega$ increases, a noticeable reduction in artifacts from the source video is observed. Concurrently, the semantic deformations in the edited output become more pronounced and accurate. For instance, the morphological structure of the dinosaur in our test case undergoes a more significant and semantically appropriate transformation with a higher $\omega$. However, our study also indicates that an excessively high guidance strength can be detrimental. When $\omega$ surpasses a certain threshold, the model begins to introduce unnatural color shifts and motion inconsistencies, which degrade the overall quality of the edited video. 

Through a systematic process of experimental analysis and evaluation, we identified $\omega = 2.75$  as the optimal value. This specific setting strikes a balance between artifact suppression and meaningful semantic deformation. At this guidance strength, the model effectively eliminates visual artifacts while producing edits that are semantically coherent and visually compelling, thereby yielding superior editing outcomes.

\paragraph{Effect of high-quality estimates and baseline estimates.}

We construct high-quality and baseline estimations by averaging the results of multiple inference runs, a strategy that effectively refines the output. As illustrated in Figure~\ref{fig:hqbl}, we evaluated configurations with varying ensemble sizes, specifically $(L_{HQ}, K) \in \{(2,1), (3,1), (3,2), (4,2)\}$. Our analysis reveals that as the number of averaging iterations for both estimations increases, there are discernible improvements in the final output, particularly in aspects such as color fidelity and overall appearance. Based on these observations, we have standardized our experimental protocol to the $(3,2)$ setting. Consequently, the high-quality estimation is generated by averaging the results of three separate inference runs, while the baseline estimation is derived from the average of two of these runs.

\paragraph{Inference Acceleration Strategies.}
To further enhance the computational efficiency of FlowDirector, we explored several optimization strategies regarding the guidance mechanism. First, we investigated the necessity of applying Classifier-Free Guidance (CFG) to the source generation branch. Our empirical analysis indicates that employing CFG on the source video yields negligible perceptual differences in the final editing results compared to using standard text conditioning alone. Consequently, we adopted an asymmetric guidance strategy: we disable CFG for the source branch (utilizing only text-conditional generation) while retaining it exclusively for the target branch. This reduction effectively halves the computational load for the source velocity estimation. Furthermore, we observe that FlowDirector is compatible with orthogonal acceleration techniques designed for diffusion models. For instance, caching the CFG residual (the difference between conditional and unconditional noise predictions) for the target branch can be seamlessly integrated into our framework, providing further reductions in inference time without compromising editing performance.

\section{Limitation}
\label{sec:appendix_limitation}

Our method aims to construct a direct editing path from the source video to the target video, bypassing the inversion process, which is prone to structural loss. Since the primary driving force for this direct editing path stems from the discrepancy between the source and target texts, varying degrees of textual difference can lead to markedly different editing outcomes. This results in incomplete text replacement, which causes substantial remnants of the original video content (Figure~\ref{fig:fail_text}). For example, modifying the source prompt $c_{\text{src}}$ (\ie, ``A large brown \textbf{bear} is walking slowly across a rocky terrain in a zoo enclosure, surrounded by stone walls and scattered greenery. The camera remains fixed, capturing the \textbf{bear}'s deliberate movements.'') to an incompletely substituted target prompt $c_{\text{tar}}$ (\ie, ``A large \textbf{dinosaur} is walking slowly across a rocky terrain in a zoo enclosure, surrounded by stone walls and scattered greenery. The camera remains fixed, capturing the \textbf{bear}'s deliberate movements.'') leads to significant residual ``bear'' information in the edited video. Furthermore, we observe that the quality of the source text $c_{\text{src}}$ also substantially affects the editing results; more comprehensive source texts tend to yield better editing outcomes compared to simpler prompts.

Similarly, our method excels in structure preservation, which is evident in tasks such as significant object editing, texture replacement, object addition/deletion, or compositional tasks. However, its performance in video style transfer is relatively limited. We attribute this to a combination of its tendency towards result preservation and being less driven by textual differences.

\section{More Qualitative Results}
\label{sec:more_results}
\begin{figure*}[htbp]
  \centering
  \includegraphics[width=\textwidth]{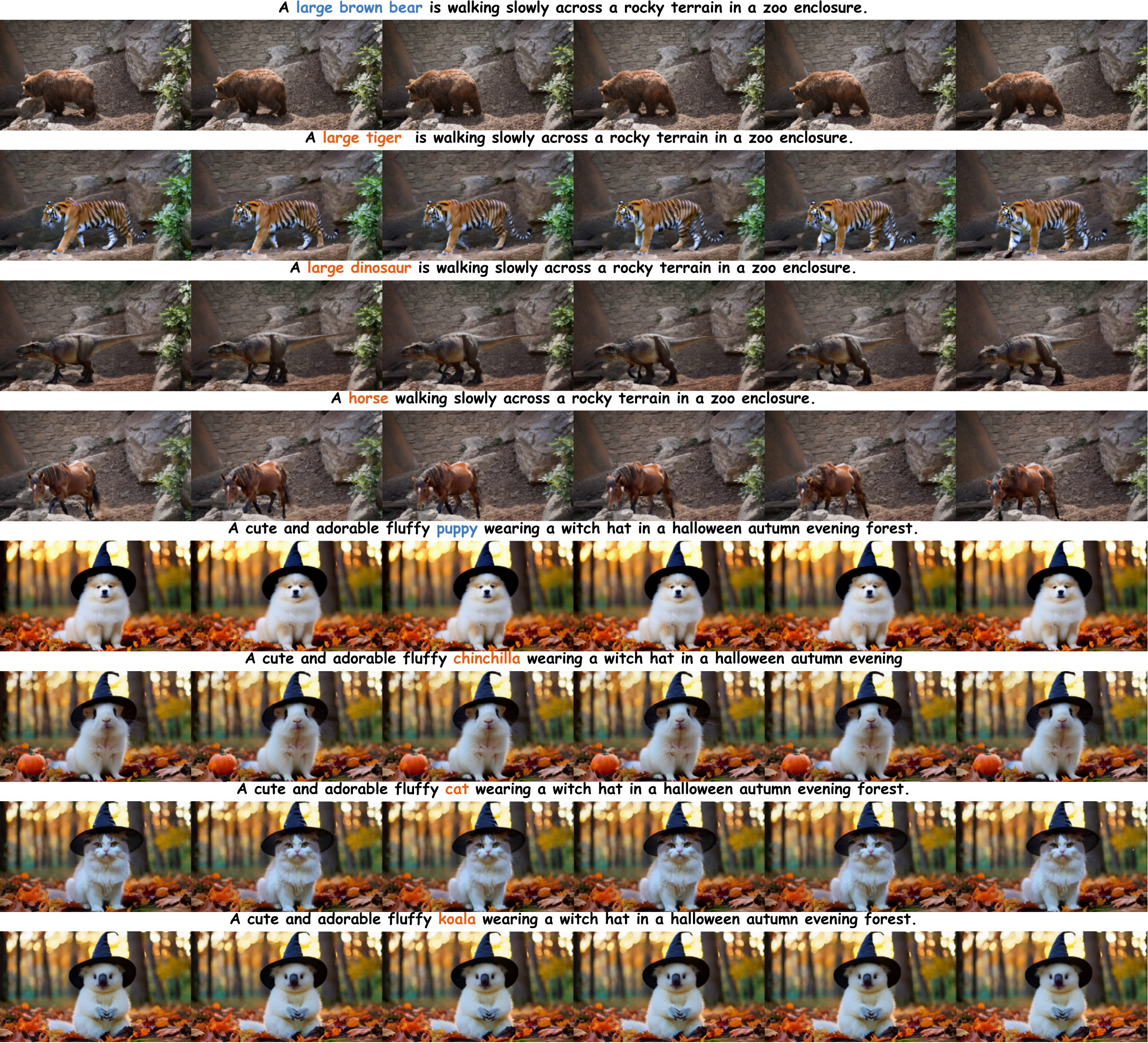}
  \caption{\textbf{More Qualitative Results.} Our method performs precise and semantically faithful edits while preserving the spatial content and motion dynamics of unedited regions. The results exhibit strong alignment with the editing instructions, high visual fidelity, and consistent temporal coherence across frames. Best viewed zoomed-in.}
  \label{fig:more_results_01}
\end{figure*}

\begin{figure*}[htbp]
  \centering
  \includegraphics[width=\textwidth]{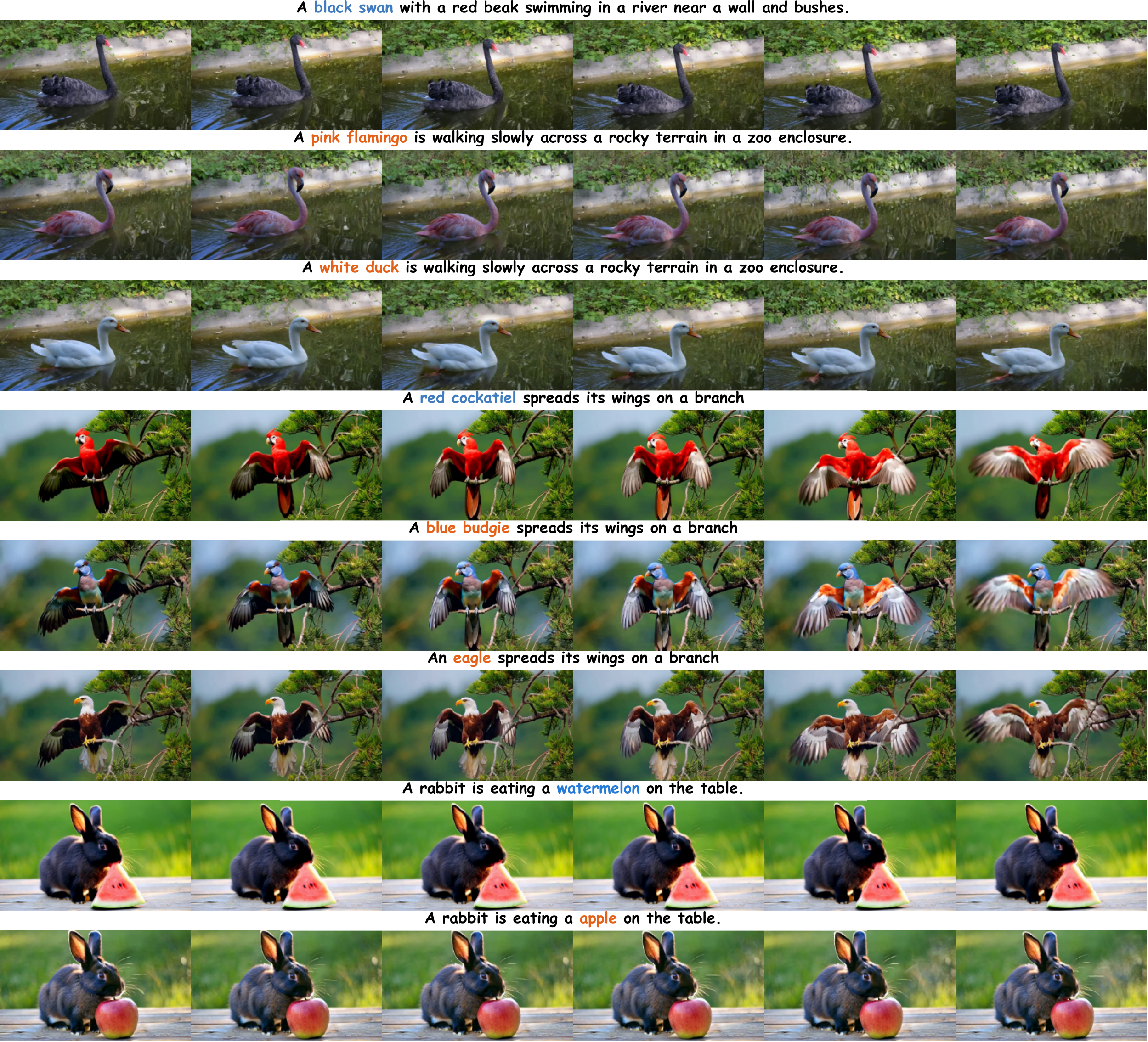}
  \caption{\textbf{More Qualitative Results.} Our method performs precise and semantically faithful edits while preserving the spatial content and motion dynamics of unedited regions. The results exhibit strong alignment with the editing instructions, high visual fidelity, and consistent temporal coherence across frames. Best viewed zoomed-in.}
\label{fig:more_results_02}
\end{figure*}

\begin{figure*}[htbp]
  \centering
  \includegraphics[width=\textwidth]{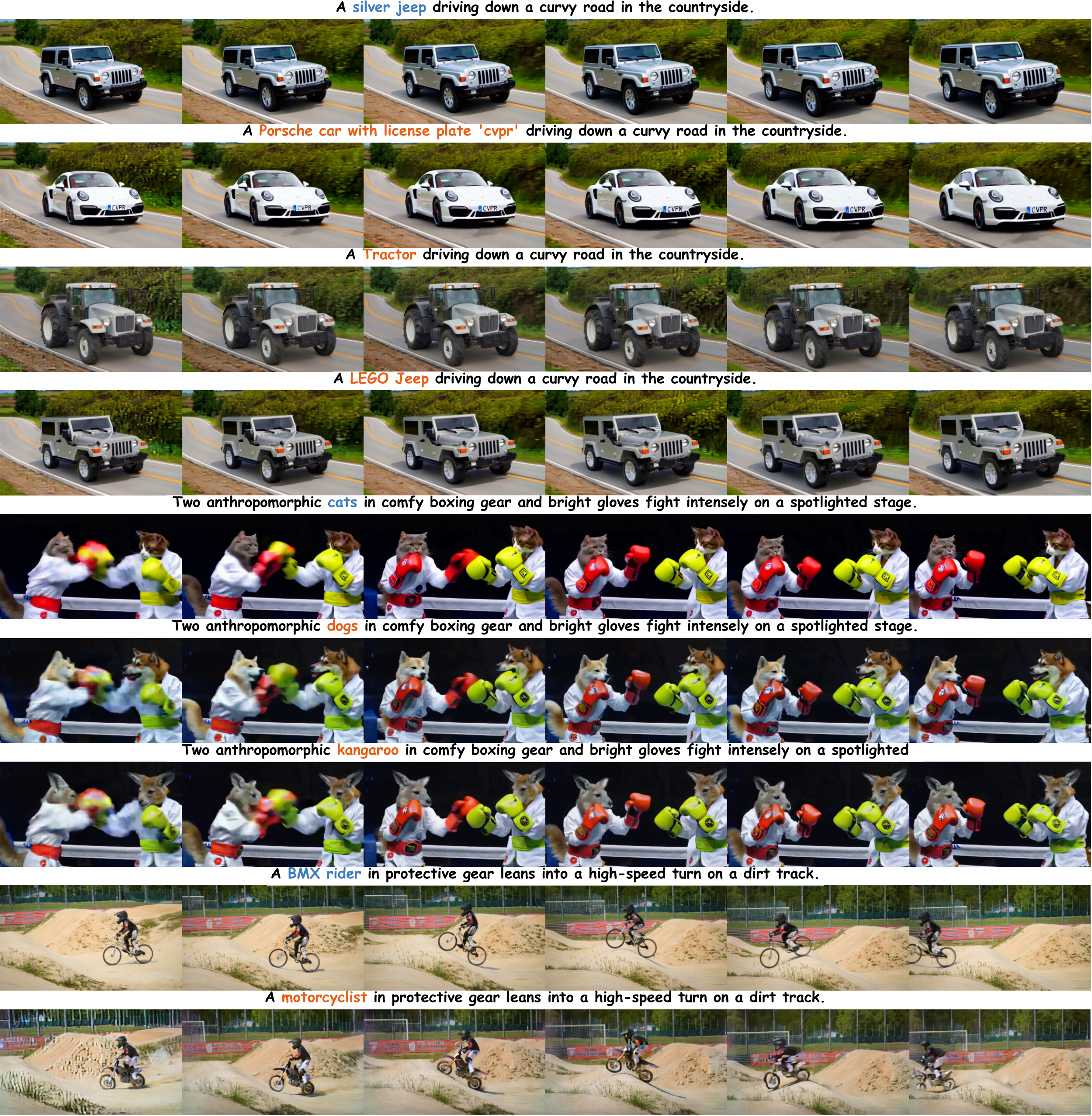}
  \caption{\textbf{More Qualitative Results.} Our method performs precise and semantically faithful edits while preserving the spatial content and motion dynamics of unedited regions. The results exhibit strong alignment with the editing instructions, high visual fidelity, and consistent temporal coherence across frames. Best viewed zoomed-in.}
  \label{fig:more_results_03}
\end{figure*}

\begin{figure*}[htbp]
  \centering
  \includegraphics[width=\textwidth]{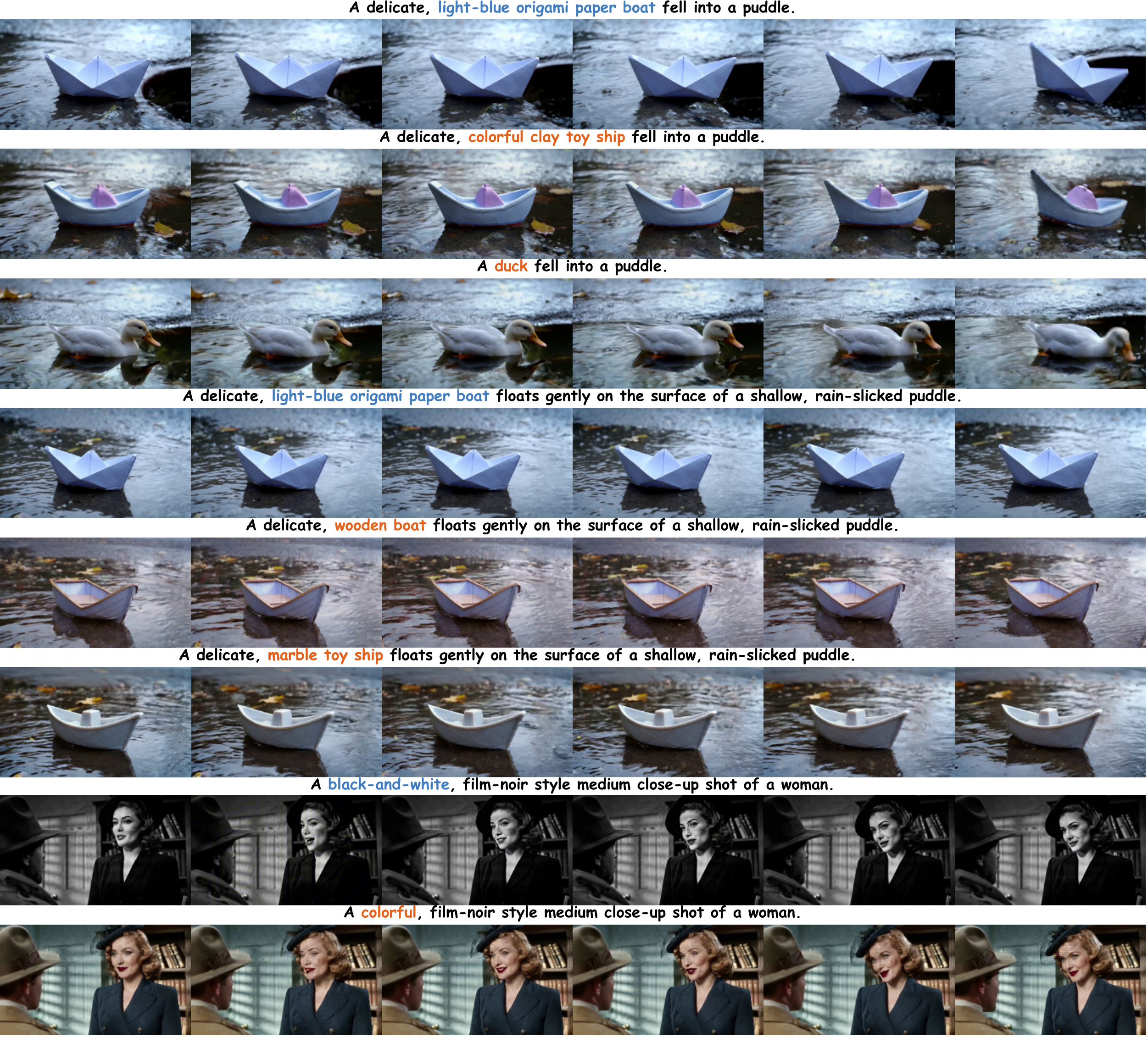}
  \caption{\textbf{More Qualitative Results.} Our method performs precise and  semantically faithful edits while preserving the spatial content and motion dynamics of unedited regions. The results exhibit strong alignment with the editing instructions, high visual fidelity, and consistent temporal coherence across frames. Best viewed zoomed-in.}
  \label{fig:more_results_04}
\end{figure*}

\begin{figure*}[htbp]
  \centering
  \includegraphics[width=\textwidth]{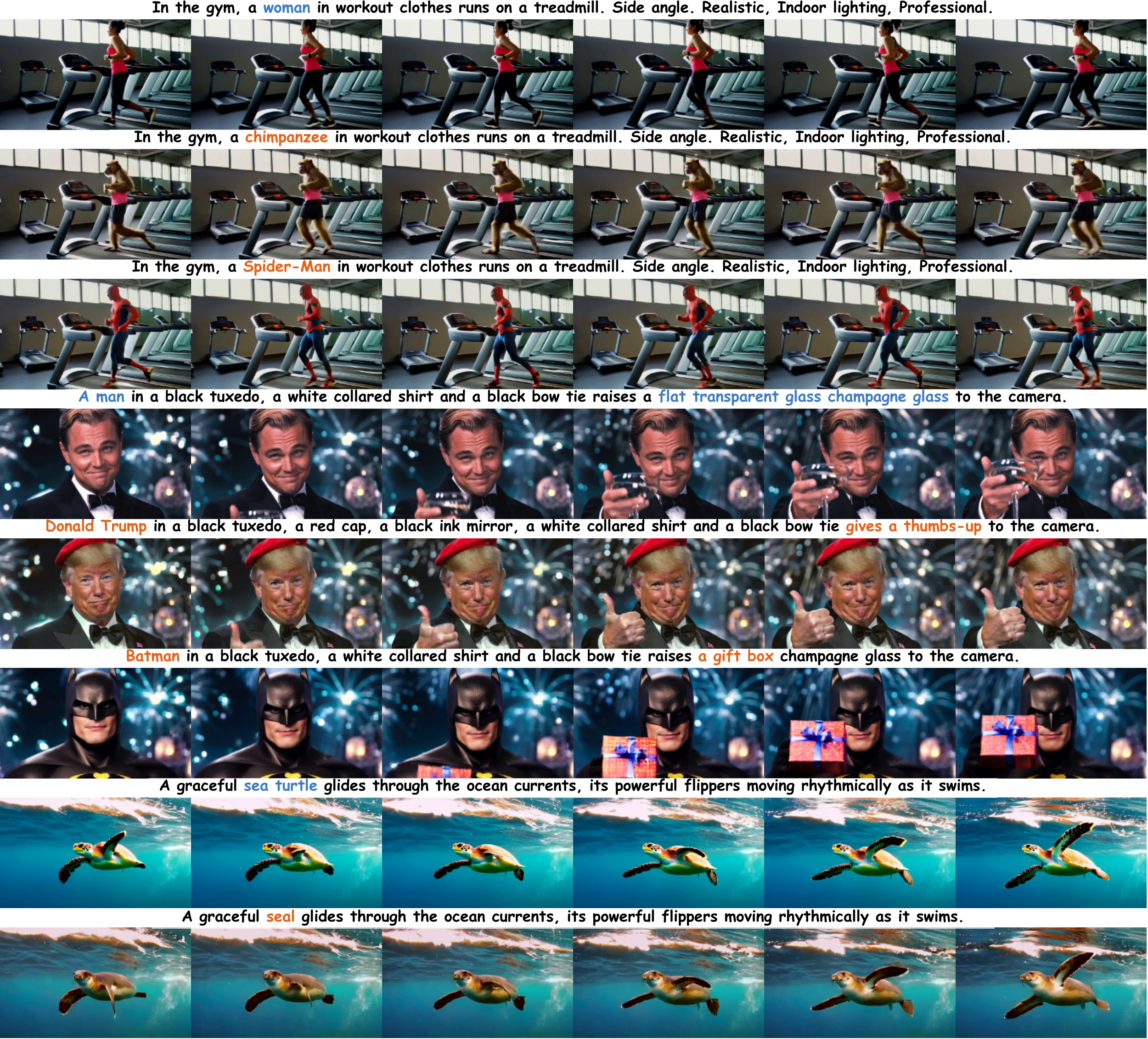}
  \caption{\textbf{More Qualitative Results.} Our method performs precise and semantically faithful edits while preserving the spatial content and motion dynamics of unedited regions. The results exhibit strong alignment with the editing instructions, high visual fidelity, and consistent temporal coherence across frames. Best viewed zoomed-in.}
  \label{fig:more_results_05}
\end{figure*}

\begin{figure*}[htbp]
  \centering
  \includegraphics[width=\textwidth]{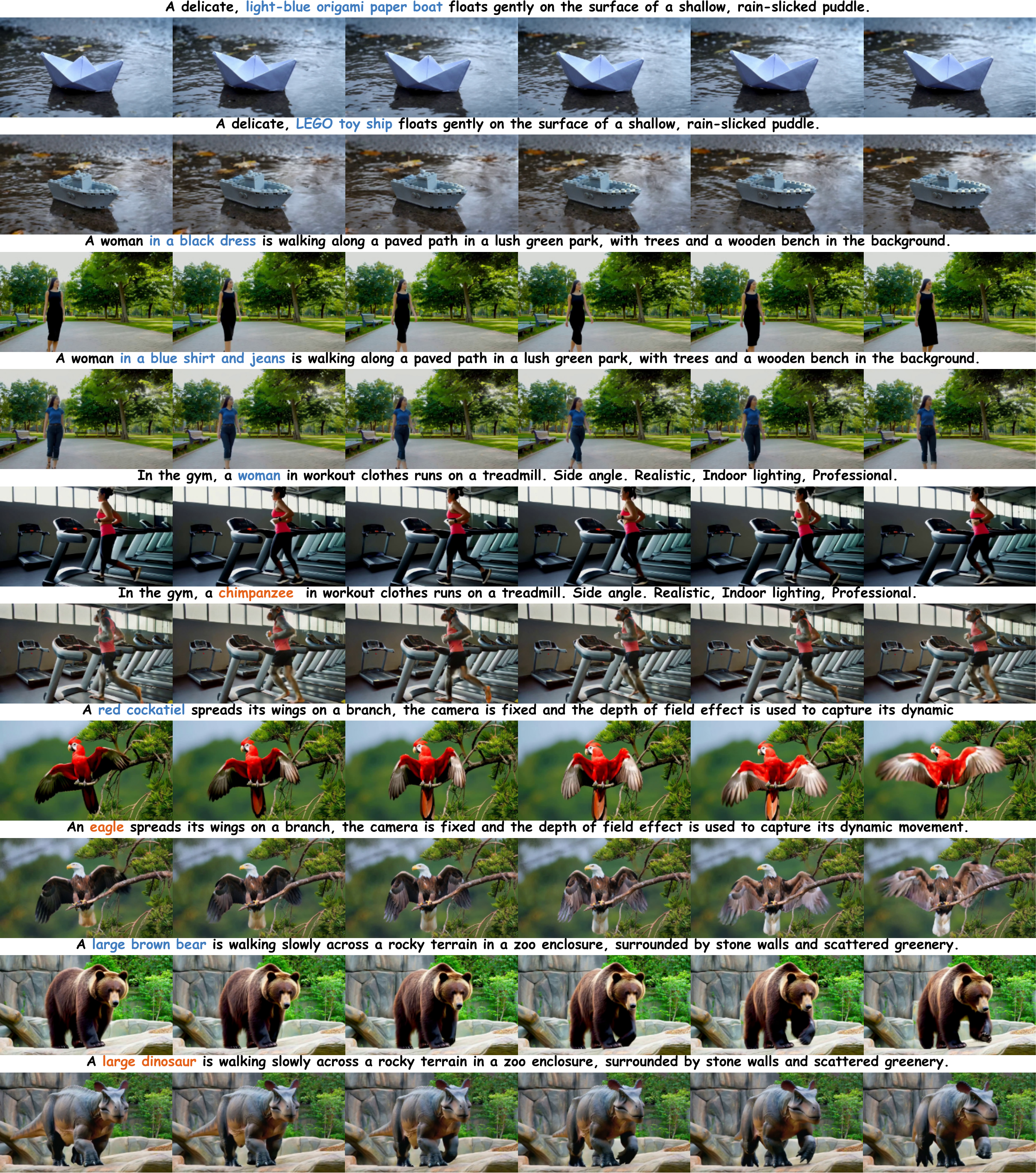}
  \caption{\textbf{Qualitative Results for Wan 14B.} Our method performs precise and semantically faithful edits while preserving the spatial content and motion dynamics of unedited regions. The results exhibit strong alignment with the editing instructions, high visual fidelity, and consistent temporal coherence across frames. Best viewed zoomed-in.}
  \label{fig:14b_show1}
\end{figure*}

In this section, we present additional qualitative results to further demonstrate the effectiveness and high quality of our video editing method. \Cref{fig:more_results_01,fig:more_results_02,fig:more_results_03,fig:more_results_04,fig:more_results_05} provide further examples of our method performing precise and semantically faithful edits while preserving the spatial content and motion dynamics of unedited regions. These results consistently exhibit strong alignment with the editing instructions, high visual fidelity, and consistent temporal coherence across frames. Figure~\ref{fig:14b_show1} also shows some editing results using Wan 2.1 14B~\cite{wan}, achieving higher editing quality and better consistency compared to the 1.3B model.

\end{document}